\documentclass[runningheads]{llncs}
\usepackage[T1]{fontenc}
\usepackage{graphicx}
\usepackage{booktabs}
\usepackage[misc]{ifsym}

\usepackage[disable,textsize=tiny]{todonotes}
\usepackage{multirow}
\usepackage{url}
\usepackage{amsmath,amsfonts,bm}

\def\eqref#1{equation~\ref{#1}}
\def\1{\bm{1}}

\def\vtheta{{\bm{\theta}}}
\def\vepsilon{{\bm{\epsilon}}}

\def\vp{{\bm{p}}}

\def\vx{{\bm{x}}}

\def\vz{{\bm{z}}}

\DeclareMathAlphabet{\mathsfit}{\encodingdefault}{\sfdefault}{m}{sl}
\SetMathAlphabet{\mathsfit}{bold}{\encodingdefault}{\sfdefault}{bx}{n}

\def\gL{{\mathcal{L}}}

\usepackage{algorithm}
\usepackage{algorithmic}
\usepackage{tikz}
\usetikzlibrary{positioning, arrows.meta, backgrounds, fit, calc, shapes.misc}

\begin{document}

\title{Reliability-Guided Adaptive Ensembling for Robust Test-Time Adaptation}

\titlerunning{Reliability-Guided Adaptive Ensembling for Robust Test-Time Adaptation}

\author{Adam Koziak\inst{1} \and 
Yuhong Guo\inst{1,2}}

\authorrunning{A.~Koziak and Y.~Guo}

\institute{School of Computer Science, Carleton University, Ottawa, Canada \and
Canada CIFAR AI Chair, Amii, Canada \\
\email{adamkoziak@cmail.carleton.ca} \quad \email{yuhong.guo@carleton.ca}}

\toctitle{Reliability-Guided Adaptive Ensembling for Robust Test-Time Adaptation}
\tocauthor{Adam Koziak, Yuhong Guo}

\maketitle

\begin{abstract}
Test-time adaptation (TTA) can mitigate domain shift without source data, but it is highly brittle under adversarially contaminated test streams, where corrupted inputs also destabilize online updates. We study robust test-time adaptation (RTTA) in the adversarial-stream setting, which remains comparatively underexplored relative to standard TTA, and propose \textbf{SAFER} (\textit{Stochastic Augmentation Framework for Enhanced Robustness}), a training-free reliability-guided augmentation wrapper for RTTA. SAFER preserves the wrapped TTA objective while replacing brittle single-view predictions with a reliability-guided pooled predictor. For each test sample, SAFER generates stochastic augmentations and aggregates their predictions through correlation-weighted pooling with outlier detection. We further study an adaptive-mixing extension that improves clean-performance retention by adjusting original-versus-augmentation weighting using feature disagreement signals. We evaluate on PACS, VLCS, and OfficeHome under PGD attacks at various attack rates. Across benchmarks, SAFER improves resilience of TTA methods to adversarial attacks while maintaining competitive clean performance.

\end{abstract}

\section{Introduction}
\label{sec:intro}

Deep models often degrade substantially when deployed on test data whose distribution differs from that seen during training. In computer vision, such domain shifts can arise from changes in appearance, acquisition conditions, corruption patterns, or broader differences between source and target environments. Test-time adaptation (TTA) has emerged as a practical response to this problem: instead of requiring source data during deployment, TTA updates a pretrained model online using only unlabeled target-domain test samples \cite{tent,tta_survey_2023,tta_survey_2024}. Representative methods such as Tent \cite{tent}, pseudo-label-based adaptation \cite{pseudolabel}, and self-distillation or feature-regularized approaches such as TSD \cite{tsd} demonstrate that test-time updates can substantially improve robustness to ordinary distribution shift without changing the original source-training pipeline.

Despite this promise, standard TTA remains fragile when the test stream is adversarially contaminated. In this setting, perturbed samples do not merely harm their own predictions: they can also corrupt the statistics, pseudo-labels, and update directions that drive online adaptation, thereby degrading predictions on future samples as well \cite{wu2023uncovering}. This failure mode is especially concerning in deployment settings where adaptation is attractive precisely because environments are changing online and source data is unavailable. While adversarial robustness has been studied extensively in supervised settings \cite{goodfellow_fgsm,pgd,croce_robustbench}, comparatively fewer works study robust test-time adaptation under adversarially perturbed online streams \cite{wu2023uncovering,tta_survey_2024}.

A straightforward response is to apply a fixed input transformation such as JPEG compression, blur, or low-pass filtering before adaptation \cite{input_transform_defense}. Such transforms can partially suppress adversarial perturbations, but they are blunt instruments: when applied uniformly, they may also erase useful information from clean samples and reduce adaptation quality. More broadly, relying on a single transformed view remains brittle, since any individual view may itself be uninformative or overly distorted. This suggests that robustness in adversarial RTTA should be framed not as trusting one view more, but as estimating which views are reliable enough to trust at all.
We therefore propose \textbf{SAFER} (\textit{Stochastic Augmentation Framework for Enhanced Robustness}), a training-free reliability-guided stochastic augmentation wrapper for robust test-time adaptation.\footnote{Code available at \url{https://github.com/AdamKoziak1/RobustTestTimeAdaptation}.} SAFER replaces brittle single-view predictions with pooled multi-augmentation predictions computed from the original input together with several stochastic augmentations. Rather than averaging these views uniformly, SAFER estimates view reliability from cross-view feature agreement, removes the least reliable view, and pools the remaining predictions with reliability-guided weights. The resulting predictor can be used as a plug-in wrapper around existing TTA methods while preserving their native adaptation objectives. In this way, SAFER targets a central weakness of online TTA under attack: over-committing adaptation to a single potentially corrupted view. Unlike generic prediction ensembling or static test-time transformation averaging~\cite{perez2021tte}, SAFER couples a \emph{specific} cross-view reliability signal and explicit outlier suppression with the wrapped method's own online update, so that the reliability-guided prediction shapes not only the current output but also \emph{future} adaptation (Fig.~\ref{fig:abl-stability}). We additionally study an adaptive-mixing extension designed to better retain clean accuracy.

Because SAFER is designed as a modular wrapper rather than a replacement adaptation algorithm, our evaluation focuses on whether the same reliability-guided predictor can consistently stabilize several representative TTA methods under attacked streams. We evaluate SAFER on PACS \cite{pacs}, VLCS \cite{vlcs}, and OfficeHome \cite{officehome} under black-box projected gradient descent (PGD) adversarial attacks. We compare against source-only anchors, standard TTA baselines, and stronger robustness- or stability-oriented baselines including EATA \cite{eata} and TeSLA \cite{tesla}, as well as simple single-transform defenses such as JPEG, blur, and FFT low-pass filtering. Across datasets and wrapped TTA methods, SAFER consistently strengthens robustness under adversarially contaminated streams while maintaining competitive clean accuracy. These results support reliability-guided stochastic ensembling as a practical and modular design principle for adversarial robust test-time adaptation.
In summary, our key contributions are as follows:
\begin{itemize}
    \item We formulate robust test-time adaptation under adversarially contaminated online test streams and position it as a comparatively underexplored setting within the broader TTA literature.
    \item We propose SAFER, a training-free wrapper that pools multi-view predictions via a reliability signal and explicit outlier dropping rather than generic ensembling, and show that this aggregation rule, not mere view multiplicity, drives the robustness gains.
    \item We evaluate the SAFER wrapper on multiple standard TTA baselines, across PACS, VLCS, and OfficeHome under black-box adversarial attacks.
\end{itemize}

\section{Related Work}
\label{sec:related}

\subsection{Test-Time Adaptation}

Test-time adaptation (TTA) addresses distribution shift by adapting a pretrained source model using unlabeled target-domain test data at deployment time, without requiring access to source samples during inference \cite{tent,tta_survey_2023,tta_survey_2024}. Compared with conventional domain adaptation, this setting is operationally attractive because it does not assume offline access to the full target set, and compared with domain generalization it allows the model to exploit target-domain information online.

Early and influential TTA methods adapt model parameters by entropy minimization on incoming batches. Tent \cite{tent}, for example, updates normalization statistics and affine parameters online by minimizing prediction entropy. Subsequent methods broadened this design space: pseudo-label-based adaptation uses model-generated labels to drive test-time updates \cite{pseudolabel}, and T3A \cite{t3a} avoids backpropagation and instead adjusts class prototypes at inference time using confident target features. TSD \cite{tsd} combines feature uniformity and feature alignment through self-distillation and local clustering, together with filtering mechanisms to reduce the impact of noisy labels. EATA \cite{eata} improves efficiency and stability by selecting informative samples and regularizing updates to reduce forgetting. TeSLA \cite{tesla} further introduces teacher-student self-learning with automatic adversarial augmentation, making it particularly relevant to our study because it already leverages augmentation during adaptation.

Our work is complementary to these methods rather than replacing their adaptation objectives. Rather than proposing yet another standalone TTA loss, we show that adversarially contaminated test streams expose a common weakness across diverse TTA algorithms, brittle reliance on single-view predictions, and address it with a plug-in prediction wrapper that strengthens multiple existing TTA objectives without modifying their native update rules.

\subsection{Robust and Stability-Oriented Test-Time Adaptation}

As TTA matured, a growing body of work began to study settings beyond benign i.i.d.\ test streams. NOTE \cite{gong2022note} highlights the failure of standard TTA under temporally correlated streams and introduces instance-aware normalization and balanced sampling to improve continual adaptation. SoTTA \cite{sotta} studies noisy test streams and proposes screening mechanisms together with parameter-stability strategies to reduce the influence of harmful samples. RoTTA \cite{rotta} targets dynamic scenarios with correlated and changing streams through robust normalization, memory-based sampling, and time-aware reweighting. MedBN \cite{medbn} focuses specifically on malicious test samples by making BN statistic estimation more robust, replacing the usual batch mean in test-time normalization with median-based statistics so that a small number of harmful samples has less influence on adaptation. 

These methods are important context, but they address robustness notions that differ from the one studied here. Their emphasis is typically on non-i.i.d.\ sampling, temporal correlation, noisy or out-of-distribution samples, or highly scarce wild adaptation regimes. By contrast, our paper focuses specifically on \emph{adversarially perturbed} online test streams, where malicious examples are injected into the target stream and can influence both current predictions and future adaptation updates. This setting remains comparatively limited within the current TTA literature \cite{wu2023uncovering,tta_survey_2024}.

The closest security-motivated evidence comes from \cite{wu2023uncovering}, which shows that TTA introduces new adversarial vulnerabilities because malicious samples can invade the adaptation dynamics of benign ones. Our work differs in emphasis: rather than primarily analyzing attacks on TTA, we develop a modular defense mechanism for adversarial robust test-time adaptation that can be wrapped around existing TTA methods.

\subsection{Transformation-Based and Multi-View Robustness Mechanisms}

A large body of adversarial-robustness work studies defenses based on adversarial training, feature denoising, or input purification and transformation \cite{pgd,xie2019featuredenoising,input_transform_defense,ferrari2023compressrestore}. However, many such approaches assume labeled source data, retraining, or deployment conditions that do not align naturally with fully online, source-free TTA. In adversarial RTTA, the defense must operate without source labels, without offline retraining, and under continual target-stream updates. Within the broader test-time adaptation literature, augmentation and multi-view reasoning have also been used to improve robustness to distribution shift. MEMO \cite{memo} adapts models using entropy minimization on the marginal prediction across augmentations, showing that multiple augmented views can provide a stronger adaptation signal than a single input alone. TeSLA \cite{tesla}, introduced above, likewise makes augmentation an explicit part of the adaptation process.

Most directly related to our setting, Perez et al.~\cite{perez2021tte} propose test-time transformation ensembling (TTE), which averages predictions over a fixed set of input transformations to improve adversarial robustness without retraining. Their approach, which we refer to as Static TTE (mean), computes a no-adaptation mean-pooled prediction over augmented views. TTE establishes that aggregating transformed views is an effective training-free robustness mechanism, but it produces a single \emph{static} robust prediction through \emph{uniform} averaging and does not perform any online model update.
Our method is related in spirit to these approaches, but differs in both mechanism and scope. SAFER does not rely on a single fixed transformation or a dedicated augmentation-trained module; it is closer to a reliability-guided multi-view inference wrapper than to a standalone transformation-based TTA objective. Relative to static transformation ensembling, SAFER differs along three axes: (i) it is a \emph{wrapper around an online TTA method} that preserves the wrapped method's native adaptation objective and update rule, rather than a one-shot static predictor; (ii) it replaces uniform transform-averaging with \emph{reliability-guided} weighting derived from cross-view feature agreement, together with explicit outlier dropping; and (iii) it targets the RTTA-specific failure mode in which contamination corrupts \emph{future} adaptation, not only the current prediction. We make this contrast quantitative in our ablations, where reliability-guided pooling (\texttt{cc\_drop}) consistently outperforms uniform \texttt{mean} pooling of the same augmentation library (Table~\ref{tab:abl-pooling-tent}).

\section{Method}

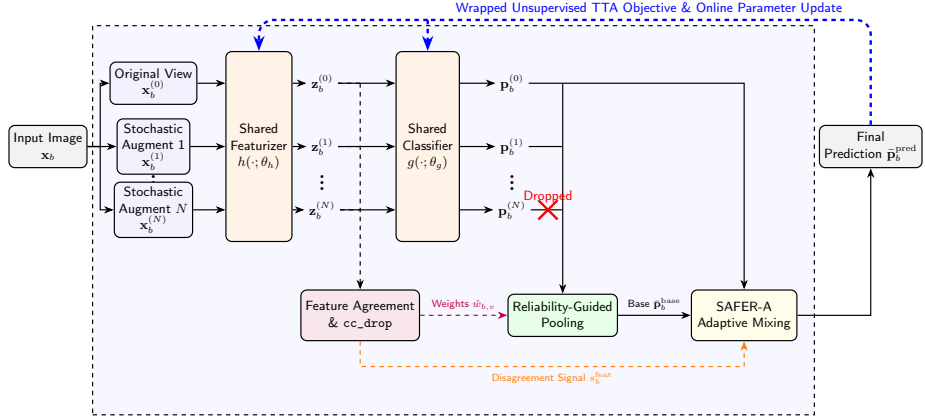
\begin{figure*}[t]
\centering
\resizebox{\textwidth}{!}{
\begin{tikzpicture}[
    >=Stealth,
    box/.style={draw, rounded corners, minimum height=1cm, align=center, font=\sffamily\small, fill=white, thick},
    model/.style={box, fill=orange!10, minimum height=4.5cm, minimum width=1.5cm},
    func/.style={box, fill=blue!5},
    safer/.style={box, fill=purple!10, minimum width=2.5cm, minimum height=1.2cm},
    highlight/.style={draw, cross out, ultra thick, red, minimum size=0.4cm, inner sep=1pt}
]

% --- 1. Input & Augmentations ---
\node[box, fill=gray!10] (input) at (0, 1.5) {Input Image\\$\mathbf{x}_b$};

\node[func] (aug0) at (2.5, 3) {Original View\\$\mathbf{x}_b^{(0)}$};
\node[func] (aug1) at (2.5, 1.5) {Stochastic\\Augment 1\\$\mathbf{x}_b^{(1)}$};
\node[font=\bfseries\Large] (dots1) at (2.5, 0.75) {$\vdots$};
\node[func] (augN) at (2.5, 0) {Stochastic\\Augment $N$\\$\mathbf{x}_b^{(N)}$};

% Arrows: Input -> Augmentations
\draw[->, thick] (input.east) -- ++(0.3,0) |- (aug0.west);
\draw[->, thick] (input.east) -- ++(0.3,0) |- (aug1.west);
\draw[->, thick] (input.east) -- ++(0.3,0) |- (augN.west);

% --- 2. Shared Featurizer & Features ---
\node[model] (feat) at (5, 1.5) {Shared\\Featurizer\\$h(\cdot; \theta_h)$};

% Arrows: Augmentations -> Featurizer
\draw[->, thick] (aug0.east) -- (aug0.east -| feat.west);
\draw[->, thick] (aug1.east) -- (aug1.east -| feat.west);
\draw[->, thick] (augN.east) -- (augN.east -| feat.west);

% Feature nodes
\node (z0) at (6.5, 3) {$\mathbf{z}_b^{(0)}$};
\node (z1) at (6.5, 1.5) {$\mathbf{z}_b^{(1)}$};
\node[font=\bfseries\Large] (dots2) at (6.5, 0.75) {$\vdots$};
\node (zN) at (6.5, 0) {$\mathbf{z}_b^{(N)}$};

% Arrows: Featurizer -> Features
\draw[->, thick] (feat.east |- z0) -- (z0.west);
\draw[->, thick] (feat.east |- z1) -- (z1.west);
\draw[->, thick] (feat.east |- zN) -- (zN.west);

% --- 3. Shared Classifier & Predictions ---
\node[model] (cls) at (9, 1.5) {Shared\\Classifier\\$g(\cdot; \theta_g)$};

% Arrows: Features -> Classifier
\draw[->, thick] (z0.east) -- (z0.east -| cls.west);
\draw[->, thick] (z1.east) -- (z1.east -| cls.west);
\draw[->, thick] (zN.east) -- (zN.east -| cls.west);

% Prediction nodes
\node (p0) at (11, 3) {$\mathbf{p}_b^{(0)}$};
\node (p1) at (11, 1.5) {$\mathbf{p}_b^{(1)}$};
\node[font=\bfseries\Large] (dots3) at (11, 0.75) {$\vdots$};
\node (pN) at (11, 0) {$\mathbf{p}_b^{(N)}$};

% Arrows: Classifier -> Predictions
\draw[->, thick] (cls.east |- p0) -- (p0.west);
\draw[->, thick] (cls.east |- p1) -- (p1.west);
\draw[->, thick] (cls.east |- pN) -- (pN.west);

% --- 4. SAFER Logic (Bottom Row) ---
\node[safer] (reli) at (7.4, -2.5) {Feature Agreement\\\& \texttt{cc\_drop}};
\node[func, fill=green!10] (pool) at (12.2, -2.5) {Reliability-Guided\\Pooling};
\node[safer, fill=yellow!10] (safera) at (16.5, -2.5) {SAFER-A\\Adaptive Mixing};

% Feature Bus dropping to Reliability
\draw[thick, dashed] (z0.east) -- (7.4, 3);
\draw[thick, dashed] (z1.east) -- (7.4, 1.5);
\draw[thick, dashed] (zN.east) -- (7.4, 0);
\draw[->, thick, dashed] (7.4, 3) -- (reli.north);

% Prediction Bus dropping to Pooling
\draw[thick] (p0.east) -- (12.2, 3);
\draw[thick] (p1.east) -- (12.2, 1.5);
\draw[thick] (pN.east) -- (12.2, 0);
\draw[->, thick] (12.2, 3) -- (pool.north);

% Dropped Outlier Visual (placed on the bus line before the pool)
\coordinate (DropPoint) at (11.85, 0);
\node[highlight] at (DropPoint) {};
\node[font=\sffamily\footnotesize, red, above=0.05cm] at (DropPoint) {Dropped};

% Routing Weights to Pool
\draw[->, thick, purple, dashed] (reli.east) -- node[above, font=\sffamily\scriptsize] {Weights $\hat{w}_{b,v}$} (pool.west);

% P0 routing to SAFER-A
\draw[->, thick] (12.2, 3) -| (safera.north);

% Pool to SAFER-A (Expanded spacing here)
\draw[->, thick] (pool.east) -- node[above, font=\sffamily\scriptsize] {Base $\bar{\mathbf{p}}_b^{\mathrm{base}}$} (safera.west);

% Disagreement Signal
\draw[->, thick, dashed, orange] (reli.south) -- ++(0,-0.6) -| node[pos=0.25, below, font=\sffamily\scriptsize, name=sig_text] {Disagreement Signal $s_b^{\mathrm{feat}}$} (safera.south);

% --- 5. Output & Feedback Loop ---
\node[box, fill=gray!10] (output) at (19.5, 1.5) {Final\\Prediction $\bar{\mathbf{p}}_b^{\mathrm{pred}}$};

% SAFER-A to Output
\draw[->, thick] (safera.east) -| (output.south);

% Top Update Loop Route (Extended to 19.5)
\draw[ultra thick, blue, dashed, rounded corners=10pt] (output.north) -- (19.5, 4.5) -- (9, 4.5) node[midway, above, font=\sffamily\small] {Wrapped Unsupervised TTA Objective \& Online Parameter Update};
\draw[->, ultra thick, blue, dashed, rounded corners=10pt] (9, 4.5) -- (cls.north);
\draw[->, ultra thick, blue, dashed, rounded corners=10pt] (9, 4.5) -- (5, 4.5) -- (feat.north);

% --- Background Wrapper Box ---
\begin{scope}[on background layer]
    \node[draw, dashed, thick, fill=blue!3, inner xsep=0.4cm, inner ysep=0.6cm, fit=(aug0) (augN) (feat) (cls) (safera) (reli) (sig_text)] {};
\end{scope}

\end{tikzpicture}
}
\caption{Overview of the SAFER framework. For each incoming test batch, the input is expanded into $N$ stochastic augmented views alongside the original view. A shared featurizer extracts representations used to compute cross-view feature agreement. The least reliable view is discarded (\texttt{cc\_drop}), and the remaining views pass through the classifier to form a reliability-weighted base prediction. The optional SAFER-A module then adaptively mixes this base prediction with the original prediction according to a feature disagreement signal. The final pooled prediction is fed into the wrapped TTA method's native objective to update the network online.}
\label{fig:safer-diagram}
\end{figure*}

We study robust test-time adaptation for a pretrained source-domain image classifier deployed on an unseen target domain, where unlabeled target samples arrive sequentially in mini-batches and a subset of the stream may be adversarially perturbed. The adapting model has no access to source data or target labels at test time. We focus on a black-box setting in which adversarial examples are prepared offline on a surrogate model and injected into the online test stream; we detail this threat model and the injection protocol in Sec.~\ref{sec:experimental-settings}. Our goal is to improve the robustness of existing TTA algorithms in this setting without changing their native adaptation objectives. For a model $f$, we denote the featurizer $h$, the prediction head $g$ such that $\vz = h(\vx; \vtheta_h)$ denotes the feature representation, and $\vp = g(\vz; \vtheta_g)$ denotes the class-probability prediction.
Standard test-time adaptation typically updates the model from a single prediction per input using an unsupervised surrogate objective such as entropy minimization. Under adversarial contamination this is brittle: a corrupted input can both induce a wrong prediction and steer subsequent online updates in a harmful direction, while a uniformly applied suppressive transform can reduce perturbations but also damage clean inputs.

SAFER is designed around this tradeoff (Fig.~\ref{fig:safer-diagram}). For each sample, it constructs multiple stochastic augmentations in addition to the original view, estimates cross-view reliability from feature agreement, removes the least reliable view, and pools the remaining predictions into a robust predictor. This reduces reliance on any single possibly corrupted view while preserving compatibility with the wrapped TTA objective; we examine the underlying assumption, including the case of multiple corrupted views, in the \emph{Assumptions and Design Rationale} discussion that closes Sec.~\ref{sec:pooling}. We additionally study an adaptive-mixing variant, SAFER-A, for applications that prioritize clean-stream retention.

\subsection{Stochastic Augmentation Construction}
\label{sec:augmentation}
For each incoming test sample $\vx_b$, SAFER constructs a set of $V=N+1$ augmentations, where $N$ is the number of stochastic augmentations and index $0$ is always the unaltered original:
\begin{equation}
\label{eq:augmentation-set}
\vx_b^{(0)}=\vx_b,\qquad
\vx_b^{(v)} = T_{\omega_{b,v}}(\vx_b),\;\; v=1,\dots,N,
\end{equation}where $T_{\omega_{b,v}}$ denotes a stochastic transformation parameterized by random variable $\omega_{b,v}$. Each view is processed by the same featurizer and classifier, producing a feature representation $\vz_b^{(v)}$ and class-probability prediction $\vp_b^{(v)}$. The transformation family is chosen to induce diverse perturbation-suppressing and appearance-altering views, so that agreement across views provides a signal of prediction reliability. Concretely, the family draws from smoothing and frequency operators (Gaussian blur, FFT low-pass filtering), additive Gaussian noise, photometric transforms, and lightweight geometric transforms. Each sampled pipeline applies at most three operators, and one must be a smoothing or frequency operator, so that every view attenuates high-frequency adversarial artifacts. Full operator definitions and parameter ranges are deferred to the supplementary material.

\subsection{Reliability-Guided Pooling}
\label{sec:pooling}
Intuitively, views that produce feature representations more consistent with the other views are treated as more reliable, while views that disagree strongly with the rest are downweighted or removed. Let $\vp_b^{(v)}\in\Delta^{K-1}$ and $\vz_b^{(v)}\in\mathbb{R}^d$ denote the class-probability prediction and feature vector of view $v$ for sample $b$. SAFER computes nonnegative reliability weights $\hat{\mathbf{w}}=(\hat w_0,\dots,\hat w_N)$ over views and forms the base pooled prediction
\begin{equation}
\label{eq:base-pooling}
\bar{\vp}_b^{\mathrm{base}}=\sum_{v=0}^{N}\hat w_v\,\vp_b^{(v)},
\qquad
\sum_{v=0}^{N}\hat w_v=1,
\qquad
\hat w_v\ge 0.
\end{equation}

The weights are derived from cross-view feature agreement. For each view $v$, we compute batch-standardized features
\begin{equation}
\label{eq:cc-standardized}
\tilde z_{b,k}^{(v)}=
\frac{z_{b,k}^{(v)}-\mu_k^{(v)}}{\sigma_k^{(v)}},
\end{equation}
where $\mu_k^{(v)}$ and $\sigma_k^{(v)}$ denote the mean and standard deviation of feature dimension $k$ over the batch of size $B$ for view $v$. We then define the reliability of view $v$ by its average agreement with the remaining views:
\begin{equation}
\label{eq:cc-reliability}
r_v=
\frac{1}{N}
\sum_{\substack{u=0\\u\neq v}}^{N}
\left[
\frac{1}{d}\sum_{k=1}^{d}
\left(
\frac{1}{B}\sum_{b=1}^{B}\tilde z_{b,k}^{(v)}\tilde z_{b,k}^{(u)}
\right)
\right].
\end{equation}

In words, $r_v$ is simply the average feature cross-correlation between view $v$ and every other view: it is high when view $v$ ``sees the same thing'' as the rest of the ensemble, and low (or negative) when it disagrees. To improve robustness to anomalous views, SAFER clips negative reliabilities, removes the least reliable view, and renormalizes:
\begin{equation}
\label{eq:cc-drop1}
\rho_v=\max(r_v,0),
\qquad
m=\arg\min_v \rho_v,
\end{equation}
\begin{equation}
\label{eq:cc-drop}
\tilde\rho_v=
\begin{cases}
0, & v=m,\\
\rho_v, & v\neq m,
\end{cases}
\qquad
\hat w_v=\frac{\tilde\rho_v}{\sum_{u=0}^{N}\tilde\rho_u}.
\end{equation}
This yields a reliability-weighted pooled predictor with lightweight outlier suppression. The removed view may be either the original view or an augmented view, depending on which is least consistent with the remainder of the set. For brevity, this cross-correlation weighting scheme with outlier dropping is referred to as \texttt{cc\_drop}.

\begin{figure}[t]
\centering
\includegraphics[alt={Five image panels of the same carousel-horse photo from the PACS Art domain. From left to right: the clean original, predicted horse with full confidence; the adversarial input at L-infinity budget 8/255, misclassified as dog and given the lowest cross-view reliability score (0.368), which the reliability-guided drop step removes; and three augmented views, all predicted horse, with higher reliability scores of about 0.67, 0.67, and 0.59. The corrupted view is the clear reliability outlier while the augmented views agree on the correct label.},width=\linewidth]{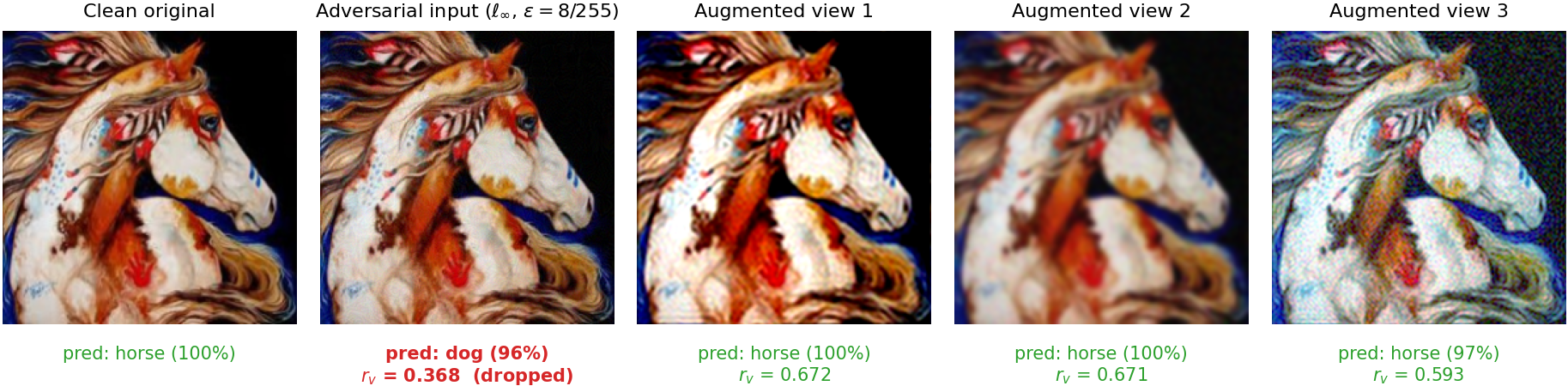}
\caption{Per-view predictions and cross-view reliability $r_v$ (Eq.~\ref{eq:cc-reliability}) for a single attacked sample; \texttt{cc\_drop} removes the adversarial input as the lowest-$r_v$ view, while the higher-$r_v$ augmented views also recover the correct label (\texttt{horse}).}
\label{fig:qualitative}
\end{figure}

\paragraph{\bf Assumptions and Design Rationale.}
Because each augmented view applies at least one smoothing or frequency operator (Sec.~\ref{sec:augmentation}), the perturbation is attenuated in every augmentation, leaving the un-augmented original as its dominant carrier. The working assumption is therefore \emph{not} that ``few of many independent samples are attacked,'' but that augmentation concentrates the surviving perturbation in one identifiable view, which then exhibits low cross-view agreement (Eq.~\ref{eq:cc-reliability}) and is flagged as a feature-space outlier while the smoothed views agree on the underlying semantics. This is also why we drop \emph{exactly one} view: removing the single least-agreeing view discards this dominant carrier while preserving ensemble diversity. The intuition is supported empirically, as dropping the worst view (\texttt{cc}~$\rightarrow$~\texttt{cc\_drop}) yields a further accuracy gain at the highest attack rate (Table~\ref{tab:abl-pooling-tent}, Sec.~\ref{sec:ablations}), and Fig.~\ref{fig:qualitative} grounds the mechanism in a concrete attacked sample, showing the per-view reliability scores $r_v$ and the view that \texttt{cc\_drop} removes.

The method does \emph{not} hinge on exactly one view being corrupted: although only one view is hard-dropped, the clipped-and-renormalized weights (Eq.~\ref{eq:cc-drop}) continue to softly down-weight every remaining low-agreement view, so additional contaminated views are still suppressed, giving graceful degradation rather than sharp failure. Clipping negative reliabilities (Eq.~\ref{eq:cc-drop1}) is deliberate: a negative mean cross-correlation marks a view as \emph{anti-correlated} with the consensus and uninformative, so clipping keeps it from distorting the normalized weights.

\subsection{Adaptive Clean-Retention Extension}
\label{sec:adaptive-alpha}
By default, the SAFER predictor returns the base pooled prediction $\bar{\vp}_b^{\mathrm{base}}$ directly, a deliberately robustness-oriented choice. We further study an extension, denoted \textbf{SAFER-A}, that instead adaptively mixes original-view mass with augmentation mass to better retain clean accuracy when the original view appears reliable, that is, when cross-view disagreement is low. SAFER-A is thus a secondary clean-retention variant; default SAFER remains the primary configuration when adversarial robustness is the goal.

We quantify disagreement between the original view and the augmented ensemble using feature-space discrepancy:
\begin{equation}
\label{eq:alpha-signal}
s_b^{\mathrm{feat}}
=
1-\cos\!\left(
\vz_b^{(0)},
\frac{1}{N}\sum_{v=1}^{N}\vz_b^{(v)}
\right),
\end{equation}

A larger value of $s_b^{\mathrm{feat}}$ indicates greater disagreement between the original view and its augmentations. This signal is mapped to a mixing coefficient through
\begin{equation}
\label{eq:alpha-sigmoid}
q_b = \sigma\!\left(\kappa(s_b^{\mathrm{feat}}-\tau)\right),
\qquad
\alpha_b = \alpha_{\mathrm{clean}} + q_b\left(\alpha_{\mathrm{atk}}-\alpha_{\mathrm{clean}}\right),
\end{equation}
where $\sigma(\cdot)$ denotes the logistic sigmoid function. $\alpha_{\mathrm{atk}}<\alpha_{\mathrm{clean}}$ are hyperparameters bounding the range of $\alpha_b$, which is the weight assigned to the original prediction relative to the augmented ensemble, with larger disagreement giving smaller weight. The final prediction weights are
\begin{equation}
\label{eq:alpha-reweight}
\tilde w_{b,0} = \alpha_b,
\qquad
\tilde w_{b,v} =
(1-\alpha_b)\,
\frac{\hat w_{b,v}}{\sum_{u=1}^{N}\hat w_{b,u}},
\quad v=1,\dots,N,
\end{equation}
yielding the final SAFER prediction
\begin{equation}
\label{eq:alpha-pred}
\bar{\vp}_b^{\mathrm{pred}}
=
\sum_{v=0}^{N}\tilde w_{b,v}\,\vp_b^{(v)}.
\end{equation}

Thus, SAFER places more weight on the augmentations when the original view disagrees strongly with augmented views, and otherwise retains greater trust in the original input.

\subsection{Plug-In Integration with Existing TTA Methods}
SAFER acts as a plug-in input wrapper for an existing TTA method. That is, SAFER changes the prediction interface seen by the wrapped method, without altering the wrapped method's underlying unsupervised objective or update rule. Denoting the final SAFER prediction by $\bar{\vp}_b^{\mathrm{pred}}$, the wrapped method optimizes its native test-time objective using $\bar{\vp}_b^{\mathrm{pred}}$ in place of the prediction from the original input alone.

For methods whose adaptation objective depends on intermediate features rather than only final predictions, SAFER analogously provides pooled features using the base reliability weights:
\begin{equation}
\label{eq:wrapper-feature-pool}
\bar{\vz}_b=\sum_{v=0}^{N}\hat w_{b,v}\,\vz_b^{(v)}.
\end{equation}

\begin{figure}[t]
\centering
\includegraphics[alt={Twin-axis line plot for Tent+SAFER on the PACS Art domain. The horizontal axis is the number of augmented views N from 1 to 4. The left vertical axis is accuracy in percent, drawn as three curves with shaded standard-deviation bands for attack rates of 0, 50, and 100 percent, all rising as N increases. The right vertical axis is measured per-step time in milliseconds, drawn as a dashed line rising roughly linearly from about 170 at N equal to 1 to about 525 at N equal to 4.},width=0.48\linewidth]{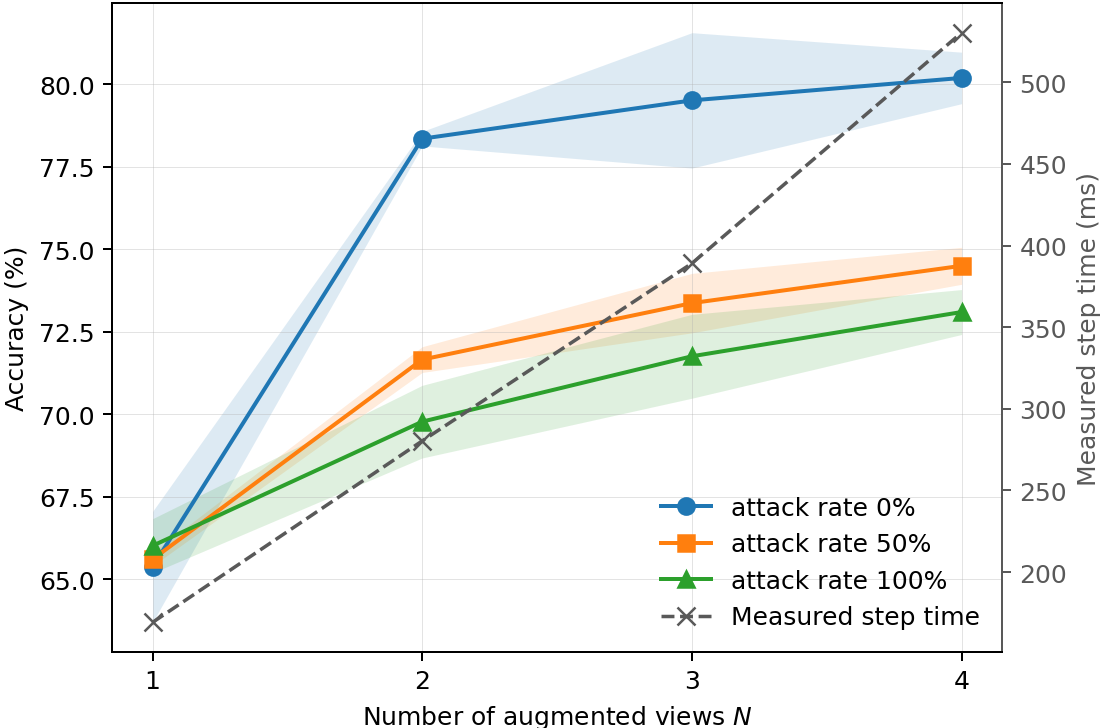}\hfill
\includegraphics[alt={Line plot of accuracy in percent versus the maximum number of augmentation operators applied per view, for Tent+SAFER on the PACS Art domain. Separate curves are shown for attack rates of 0, 50, and 100 percent.},width=0.48\linewidth]{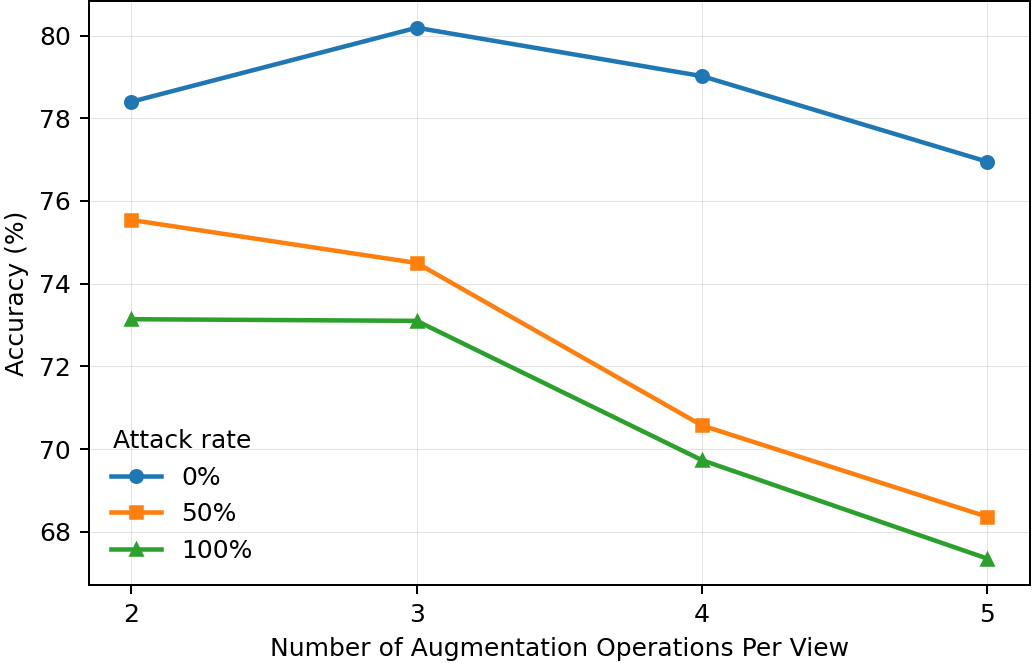}
\caption{Single-domain sensitivity plots on PACS:Art Painting for Tent+SAFER. Left: effect of the number of views. Right: effect of the number of operations per augmentation.}
\label{fig:abl-views-maxops}
\end{figure}

These pooled features use the base non-adaptive reliability weights rather than the adaptive original-view mixing of SAFER-A: adaptive mixing affects only the final prediction returned at inference time, while wrapped feature-driven objectives continue to receive the base reliability-guided pooled features. This keeps SAFER compatible with both prediction-driven and feature-driven TTA methods while preserving the wrapped algorithm's adaptation mechanism. SAFER adds no new parameters and no extra adaptation step; we analyze its per-batch overhead in Sec.~\ref{sec:experimental-settings} (Fig.~\ref{fig:abl-views-maxops}).

\section{Experiments}

\begin{table*}[t]
\centering
\caption{PACS accuracy (\%) under $\ell_\infty$ attacks with attack rates 0\% and 100\% per target domain (bold: best, underline: second best per column).}
\label{tab:pacs-main}\resizebox{\textwidth}{!}{
\setlength{\tabcolsep}{1pt}
\begin{tabular}{l|rr|rr|rr|rr}
\hline
\multirow{2}{*}{Method} & \multicolumn{2}{c|}{Art Painting} & \multicolumn{2}{c|}{Cartoon} & \multicolumn{2}{c|}{Photo} & \multicolumn{2}{c}{Sketch} \\
& 0\% & 100\% & 0\% & 100\% & 0\% & 100\% & 0\% & 100\% \\
\hline

ERM & 78.82$_{\pm 2.00}$ & 12.40$_{\pm 3.28}$ & 75.51$_{\pm 2.22}$ & 22.94$_{\pm 3.20}$ & 94.99$_{\pm 0.42}$ & 59.34$_{\pm 3.60}$ & 71.10$_{\pm 1.86}$ & 55.61$_{\pm 1.14}$ \\
Robust ERM & 47.69$_{\pm 0.93}$ & 47.23$_{\pm 0.89}$ & 64.04$_{\pm 1.47}$ & 63.77$_{\pm 1.62}$ & 62.20$_{\pm 0.62}$ & 62.08$_{\pm 0.60}$ & 58.45$_{\pm 1.94}$ & 58.29$_{\pm 1.95}$ \\

TeSLA & 82.21$_{\pm 0.91}$ & 45.38$_{\pm 4.63}$ & 80.86$_{\pm 0.42}$ & 61.43$_{\pm 1.09}$ & 95.11$_{\pm 0.09}$ & 71.32$_{\pm 3.28}$ & 74.16$_{\pm 0.60}$ & 52.93$_{\pm 3.90}$ \\
EATA & 81.61$_{\pm 0.71}$ & \phantom{0}7.39$_{\pm 1.62}$ & 81.40$_{\pm 0.73}$ & 36.08$_{\pm 3.10}$ & 95.27$_{\pm 0.51}$ & 38.00$_{\pm 4.52}$ & 73.00$_{\pm 0.91}$ & 28.54$_{\pm 4.95}$ \\
Tent & 83.72$_{\pm 0.35}$ & \phantom{0}4.98$_{\pm 1.89}$ & 83.45$_{\pm 0.79}$ & 35.62$_{\pm 5.95}$ & \underline{95.91}$_{\pm 0.66}$ & 40.82$_{\pm 5.20}$ & 78.31$_{\pm 1.15}$ & 29.91$_{\pm 6.33}$ \\
Tent$_{\text{+MedBN}}$ & 82.62$_{\pm 0.93}$ & \phantom{0}4.57$_{\pm 1.56}$ & 81.41$_{\pm 0.47}$ & 39.55$_{\pm 2.78}$ & 95.87$_{\pm 0.59}$ & 29.16$_{\pm 5.09}$ & 75.35$_{\pm 1.05}$ & 26.73$_{\pm 7.50}$ \\\hline
Tent$_{\text{+SAFER}}$ & 80.06$_{\pm 1.25}$ & \underline{71.74}$_{\pm 0.52}$ & 83.11$_{\pm 0.67}$ & 79.24$_{\pm 0.23}$ & 94.63$_{\pm 0.19}$ & \textbf{90.78}$_{\pm 0.87}$ & \underline{80.28}$_{\pm 0.92}$ & \textbf{76.33}$_{\pm 0.83}$ \\
Tent$_{\text{+SAFER-A}}$ & 82.60$_{\pm 1.17}$ & 48.44$_{\pm 1.86}$ & 83.73$_{\pm 0.77}$ & 64.68$_{\pm 0.46}$ & 95.71$_{\pm 0.46}$ & 75.27$_{\pm 1.10}$ & \textbf{81.14}$_{\pm 1.01}$ & 66.61$_{\pm 2.98}$ \\\hline
PL & 85.76$_{\pm 2.02}$ & \phantom{0}2.78$_{\pm 1.32}$ & 82.10$_{\pm 3.34}$ & 37.54$_{\pm 1.92}$ & 95.37$_{\pm 0.88}$ & 35.19$_{\pm 8.23}$ & 72.91$_{\pm 5.82}$ & 28.34$_{\pm 9.69}$ \\\hline
PL$_{\text{+SAFER}}$ & 79.13$_{\pm 3.13}$ & \textbf{72.97}$_{\pm 1.02}$ & 85.00$_{\pm 1.75}$ & \underline{79.62}$_{\pm 1.99}$ & 93.93$_{\pm 0.95}$ & 89.92$_{\pm 1.29}$ & 74.73$_{\pm 1.80}$ & 70.61$_{\pm 6.43}$ \\
PL$_{\text{+SAFER-A}}$ & 83.63$_{\pm 1.47}$ & 42.02$_{\pm 2.97}$ & 84.33$_{\pm 2.83}$ & 63.41$_{\pm 3.15}$ & 94.43$_{\pm 1.20}$ & 71.82$_{\pm 12.41}$ & 77.76$_{\pm 7.03}$ & 64.43$_{\pm 6.28}$ \\\hline
TSD & \textbf{86.90}$_{\pm 1.59}$ & \phantom{0}1.61$_{\pm 0.75}$ & \underline{88.10}$_{\pm 0.62}$ & 45.05$_{\pm 4.41}$ & \textbf{96.31}$_{\pm 0.89}$ & 45.63$_{\pm 9.06}$ & 78.90$_{\pm 3.75}$ & 32.72$_{\pm 15.87}$ \\\hline
TSD$_{\text{+SAFER}}$ & 83.32$_{\pm 1.20}$ & 71.18$_{\pm 3.13}$ & 87.57$_{\pm 1.22}$ & \textbf{82.20}$_{\pm 0.16}$ & 94.21$_{\pm 1.05}$ & \underline{90.12}$_{\pm 0.33}$ & 76.00$_{\pm 1.04}$ & \underline{71.45}$_{\pm 3.86}$ \\
TSD$_{\text{+SAFER-A}}$ & \underline{85.87}$_{\pm 1.89}$ & 56.01$_{\pm 2.93}$ & \textbf{88.41}$_{\pm 0.97}$ & 64.56$_{\pm 1.44}$ & 95.53$_{\pm 1.16}$ & 79.38$_{\pm 0.33}$ & 78.62$_{\pm 1.97}$ & 64.82$_{\pm 0.90}$ \\

\hline
\end{tabular}
}
\end{table*}

\begin{table*}[t]
\centering
\caption{VLCS accuracy (\%) under $\ell_\infty$ attacks with attack rates 0\% and 100\% per target domain (bold: best, underline: second best per column).}
\label{tab:vlcs-main}
\resizebox{\textwidth}{!}{
\setlength{\tabcolsep}{1pt}
\begin{tabular}{l|rr|rr|rr|rr}
\hline
\multirow{2}{*}{Method} & \multicolumn{2}{c|}{Caltech101} & \multicolumn{2}{c|}{LabelMe} & \multicolumn{2}{c|}{SUN09} & \multicolumn{2}{c}{VOC2007} \\
& 0\% & 100\% & 0\% & 100\% & 0\% & 100\% & 0\% & 100\% \\
\hline

ERM & \underline{95.27}$_{\pm 1.80}$ & 49.63$_{\pm 3.76}$ & 64.76$_{\pm 0.76}$ & 20.36$_{\pm 4.74}$ & \underline{68.96}$_{\pm 0.82}$ & \phantom{0}7.33$_{\pm 3.43}$ & \underline{72.27}$_{\pm 3.35}$ & \phantom{0}8.71$_{\pm 4.36}$ \\
Robust ERM & 61.51$_{\pm 0.11}$ & 61.48$_{\pm 0.12}$ & 53.61$_{\pm 0.57}$ & 53.69$_{\pm 0.52}$ & 43.70$_{\pm 0.90}$ & \underline{43.65}$_{\pm 0.78}$ & 48.68$_{\pm 0.58}$ & 48.52$_{\pm 0.54}$ \\

\hline
TeSLA & 82.54$_{\pm 3.44}$ & 63.86$_{\pm 2.59}$ & 59.60$_{\pm 1.37}$ & 40.51$_{\pm 2.13}$ & 61.84$_{\pm 0.90}$ & 36.25$_{\pm 0.32}$ & 72.02$_{\pm 1.42}$ & 46.26$_{\pm 2.62}$ \\
EATA & 82.38$_{\pm 3.02}$ & 47.94$_{\pm 1.39}$ & 59.50$_{\pm 1.32}$ & 20.39$_{\pm 3.68}$ & 61.95$_{\pm 0.98}$ & 13.84$_{\pm 3.34}$ & 71.03$_{\pm 1.57}$ & 22.10$_{\pm 2.47}$ \\
Tent & 87.59$_{\pm 3.86}$ & 53.80$_{\pm 3.46}$ & 62.44$_{\pm 0.92}$ & 15.95$_{\pm 6.02}$ & 65.99$_{\pm 1.76}$ & \phantom{0}9.18$_{\pm 5.21}$ & \textbf{72.62}$_{\pm 1.29}$ & 21.45$_{\pm 3.63}$ \\
Tent$_{\text{+MedBN}}$ & 85.84$_{\pm 2.43}$ & 50.44$_{\pm 4.33}$ & 62.09$_{\pm 1.00}$ & 19.99$_{\pm 5.75}$ & 64.04$_{\pm 0.81}$ & 11.35$_{\pm 3.42}$ & 71.38$_{\pm 2.54}$ & 20.16$_{\pm 3.08}$ \\\hline
Tent$_{\text{+SAFER}}$ & 87.02$_{\pm 4.07}$ & 83.67$_{\pm 2.94}$ & 62.42$_{\pm 0.85}$ & \underline{56.76}$_{\pm 0.87}$ & 61.08$_{\pm 1.23}$ & \textbf{48.21}$_{\pm 1.23}$ & 67.03$_{\pm 0.96}$ & \textbf{59.73}$_{\pm 0.18}$ \\
Tent$_{\text{+SAFER-A}}$ & 87.92$_{\pm 3.59}$ & 73.26$_{\pm 1.83}$ & \textbf{65.89}$_{\pm 0.59}$ & 42.32$_{\pm 6.23}$ & 64.46$_{\pm 0.63}$ & 33.99$_{\pm 3.97}$ & 67.57$_{\pm 0.88}$ & 48.84$_{\pm 1.46}$ \\
PL & 90.62$_{\pm 6.88}$ & 52.25$_{\pm 5.02}$ & \underline{65.59}$_{\pm 1.66}$ & \phantom{0}4.46$_{\pm 1.22}$ & 65.60$_{\pm 7.99}$ & \phantom{0}4.70$_{\pm 2.73}$ & 68.02$_{\pm 3.16}$ & 20.44$_{\pm 5.04}$ \\\hline
PL$_{\text{+SAFER}}$ & 87.49$_{\pm 6.25}$ & \underline{84.31}$_{\pm 5.09}$ & 61.66$_{\pm 0.64}$ & 53.82$_{\pm 1.94}$ & 50.98$_{\pm 1.08}$ & 41.30$_{\pm 2.20}$ & 64.72$_{\pm 1.98}$ & \underline{53.92}$_{\pm 2.45}$ \\
PL$_{\text{+SAFER-A}}$ & 88.81$_{\pm 6.96}$ & 72.08$_{\pm 2.29}$ & 62.46$_{\pm 2.10}$ & 45.68$_{\pm 11.73}$ & 55.13$_{\pm 6.52}$ & 36.62$_{\pm 3.83}$ & 62.99$_{\pm 0.98}$ & 51.55$_{\pm 1.04}$ \\
TSD & \textbf{95.64}$_{\pm 2.32}$ & 48.62$_{\pm 7.41}$ & 63.77$_{\pm 0.52}$ & \phantom{0}8.36$_{\pm 5.84}$ & 67.14$_{\pm 1.56}$ & \phantom{0}2.51$_{\pm 1.47}$ & 66.44$_{\pm 6.11}$ & 13.57$_{\pm 3.83}$ \\\hline
TSD$_{\text{+SAFER}}$ & 94.61$_{\pm 1.38}$ & \textbf{84.71}$_{\pm 6.21}$ & 64.12$_{\pm 0.94}$ & \textbf{56.94}$_{\pm 4.91}$ & 66.72$_{\pm 2.74}$ & 42.92$_{\pm 1.47}$ & 58.51$_{\pm 3.83}$ & 48.30$_{\pm 1.14}$ \\
TSD$_{\text{+SAFER-A}}$ & 94.49$_{\pm 1.66}$ & 75.81$_{\pm 1.06}$ & 63.83$_{\pm 0.71}$ & 51.51$_{\pm 0.66}$ & \textbf{69.37}$_{\pm 1.95}$ & 36.23$_{\pm 1.16}$ & 60.20$_{\pm 2.08}$ & 48.75$_{\pm 2.35}$ \\

\hline
\end{tabular}
}
\end{table*}

\begin{table*}[t]
\centering
\caption{OfficeHome accuracy (\%) under $\ell_\infty$ attacks with attack rates 0\% and 100\% per target domain (bold: best, underline: second best per column).}
\label{tab:officehome-main}
\resizebox{\textwidth}{!}{
\setlength{\tabcolsep}{1pt}
\begin{tabular}{l|rr|rr|rr|rr}
\hline
\multirow{2}{*}{Method} & \multicolumn{2}{c|}{Art} & \multicolumn{2}{c|}{Clipart} & \multicolumn{2}{c|}{Product} & \multicolumn{2}{c}{RealWorld} \\
& 0\% & 100\% & 0\% & 100\% & 0\% & 100\% & 0\% & 100\% \\
\hline

ERM & 56.15$_{\pm 0.27}$ & \phantom{0}2.94$_{\pm 0.37}$ & 49.00$_{\pm 0.19}$ & \phantom{0}8.71$_{\pm 0.94}$ & \underline{71.50}$_{\pm 0.62}$ & 13.29$_{\pm 0.47}$ & \textbf{73.77}$_{\pm 0.68}$ & \phantom{0}4.32$_{\pm 0.26}$ \\
Robust ERM & 28.80$_{\pm 1.15}$ & 28.58$_{\pm 1.18}$ & 35.54$_{\pm 0.36}$ & 35.57$_{\pm 0.37}$ & 43.20$_{\pm 0.36}$ & 43.08$_{\pm 0.25}$ & 44.43$_{\pm 0.54}$ & 44.28$_{\pm 0.53}$ \\

\hline
TeSLA & 55.49$_{\pm 0.75}$ & 26.15$_{\pm 2.81}$ & 47.28$_{\pm 0.70}$ & 29.37$_{\pm 2.38}$ & 70.26$_{\pm 0.18}$ & 48.36$_{\pm 3.27}$ & 72.61$_{\pm 0.44}$ & 41.06$_{\pm 2.28}$ \\
EATA & 54.95$_{\pm 0.78}$ & \phantom{0}3.98$_{\pm 0.31}$ & 49.60$_{\pm 0.48}$ & 11.96$_{\pm 1.05}$ & 70.50$_{\pm 0.14}$ & 20.57$_{\pm 1.26}$ & 72.91$_{\pm 0.21}$ & \phantom{0}7.73$_{\pm 0.61}$ \\
Tent & \textbf{56.43}$_{\pm 1.14}$ & \phantom{0}2.62$_{\pm 0.36}$ & \underline{51.43}$_{\pm 0.34}$ & 11.66$_{\pm 2.15}$ & \textbf{71.70}$_{\pm 0.49}$ & 19.61$_{\pm 1.19}$ & \underline{73.36}$_{\pm 0.52}$ & \phantom{0}4.04$_{\pm 0.27}$ \\
Tent$_{\text{+MedBN}}$ & 54.66$_{\pm 1.55}$ & \phantom{0}2.84$_{\pm 0.39}$ & 47.96$_{\pm 0.33}$ & 14.24$_{\pm 0.69}$ & 67.58$_{\pm 0.11}$ & 21.21$_{\pm 1.56}$ & 72.10$_{\pm 0.54}$ & \phantom{0}5.35$_{\pm 0.48}$ \\\hline
Tent$_{\text{+SAFER}}$ & 51.16$_{\pm 0.87}$ & \underline{40.26}$_{\pm 0.33}$ & 51.04$_{\pm 0.71}$ & \textbf{46.10}$_{\pm 1.06}$ & 67.30$_{\pm 0.72}$ & \textbf{60.03}$_{\pm 0.45}$ & 68.63$_{\pm 0.31}$ & \textbf{60.55}$_{\pm 0.73}$ \\
Tent$_{\text{+SAFER-A}}$ & 54.73$_{\pm 1.11}$ & 19.91$_{\pm 1.23}$ & \textbf{51.83}$_{\pm 0.82}$ & 32.01$_{\pm 1.49}$ & 70.18$_{\pm 0.70}$ & 44.81$_{\pm 0.77}$ & 71.13$_{\pm 0.12}$ & 38.10$_{\pm 0.76}$ \\
PL & 51.22$_{\pm 1.20}$ & \phantom{0}3.15$_{\pm 0.43}$ & 39.12$_{\pm 1.74}$ & 11.42$_{\pm 0.57}$ & 65.80$_{\pm 1.12}$ & 14.89$_{\pm 4.81}$ & 65.70$_{\pm 0.95}$ & \phantom{0}3.37$_{\pm 0.56}$ \\\hline
PL$_{\text{+SAFER}}$ & 44.14$_{\pm 1.34}$ & 30.56$_{\pm 0.31}$ & 33.47$_{\pm 2.16}$ & 30.10$_{\pm 2.91}$ & 56.60$_{\pm 1.06}$ & 45.96$_{\pm 0.32}$ & 55.95$_{\pm 2.23}$ & 44.42$_{\pm 0.69}$ \\
PL$_{\text{+SAFER-A}}$ & 47.05$_{\pm 0.93}$ & 15.89$_{\pm 0.65}$ & 36.77$_{\pm 5.85}$ & 20.66$_{\pm 0.94}$ & 60.08$_{\pm 0.97}$ & 29.38$_{\pm 2.27}$ & 59.54$_{\pm 2.86}$ & 20.67$_{\pm 3.59}$ \\
TSD & \underline{56.32}$_{\pm 0.51}$ & \phantom{0}2.51$_{\pm 0.51}$ & 48.10$_{\pm 1.60}$ & 12.09$_{\pm 2.38}$ & 69.56$_{\pm 1.20}$ & 19.32$_{\pm 1.28}$ & 69.42$_{\pm 0.57}$ & \phantom{0}3.03$_{\pm 0.50}$ \\\hline
TSD$_{\text{+SAFER}}$ & 51.41$_{\pm 0.99}$ & \textbf{40.98}$_{\pm 1.04}$ & 44.50$_{\pm 0.93}$ & \underline{38.06}$_{\pm 1.37}$ & 63.26$_{\pm 1.89}$ & \underline{54.01}$_{\pm 0.68}$ & 63.06$_{\pm 1.14}$ & \underline{48.98}$_{\pm 1.95}$ \\
TSD$_{\text{+SAFER-A}}$ & 54.58$_{\pm 1.09}$ & 20.31$_{\pm 0.33}$ & 46.09$_{\pm 1.40}$ & 30.44$_{\pm 1.01}$ & 66.33$_{\pm 2.06}$ & 42.73$_{\pm 0.35}$ & 65.71$_{\pm 0.86}$ & 32.64$_{\pm 2.62}$ \\

\hline
\end{tabular}
}
\end{table*}

\begin{table}[t]
\centering
\caption{Single-transform defenses and SAFER pooling rules on Tent (PACS average).}
\label{tab:abl-pooling-tent}
\begin{tabular}{lccc}
\hline
Method & 0\% & 50\% & 100\%  \\
\hline
Tent & 85.35 & 53.08 & 27.83 \\
Tent + JPEG & 84.59 & 80.12 & 75.82 \\
Tent + FFT & 83.55 & 73.23 & 65.86 \\
Tent + Blur & 85.05 & 58.52 & 37.60 \\
\hline
Tent + SAFER (mean) & 84.34 & 80.39 & 76.78 \\
Tent + SAFER (entropy) & 84.58 & 80.16 & 76.18 \\
Tent + SAFER (cc) & 84.46 & 80.63 & 78.36  \\
Tent + SAFER (cc\_drop) & 84.55 & 80.75 & 79.83  \\
\hline
\end{tabular}
\end{table}

\begin{table}[t]
\centering
\caption{Alpha-mode ablation for Tent + SAFER, accuracy over the Art Painting domain of PACS.}
\label{tab:abl-alpha-mode}
\begin{tabular}{lccc}
\hline
Method & 0\% & 50\% & 100\%  \\
\hline
Tent & 83.72 & 44.78 & 4.98 \\
Tent + SAFER & 80.19 & 74.50 & 73.10 \\
Tent + SAFER-A (Step) & 82.57 & 72.09 & 61.15  \\
Tent + SAFER-A (Linear) & 82.75 & 71.73 & 59.68  \\
Tent + SAFER-A (Sigmoid) & 83.54 & 66.81 & 49.87 \\
\hline
\end{tabular}
\end{table}

\begin{figure}[t]
\centering
\includegraphics[alt={Line plot of rolling batch accuracy in percent, using a 16-batch moving window, across the test stream under a fully attacked stream on the PACS Art domain. Curves are shown for Tent, Tent+SAFER, Tent+SAFER-A, and a no-adaptation Static TTE mean reference. Tent is low and unstable while the SAFER variants stay high and stable.},width=0.48\linewidth]{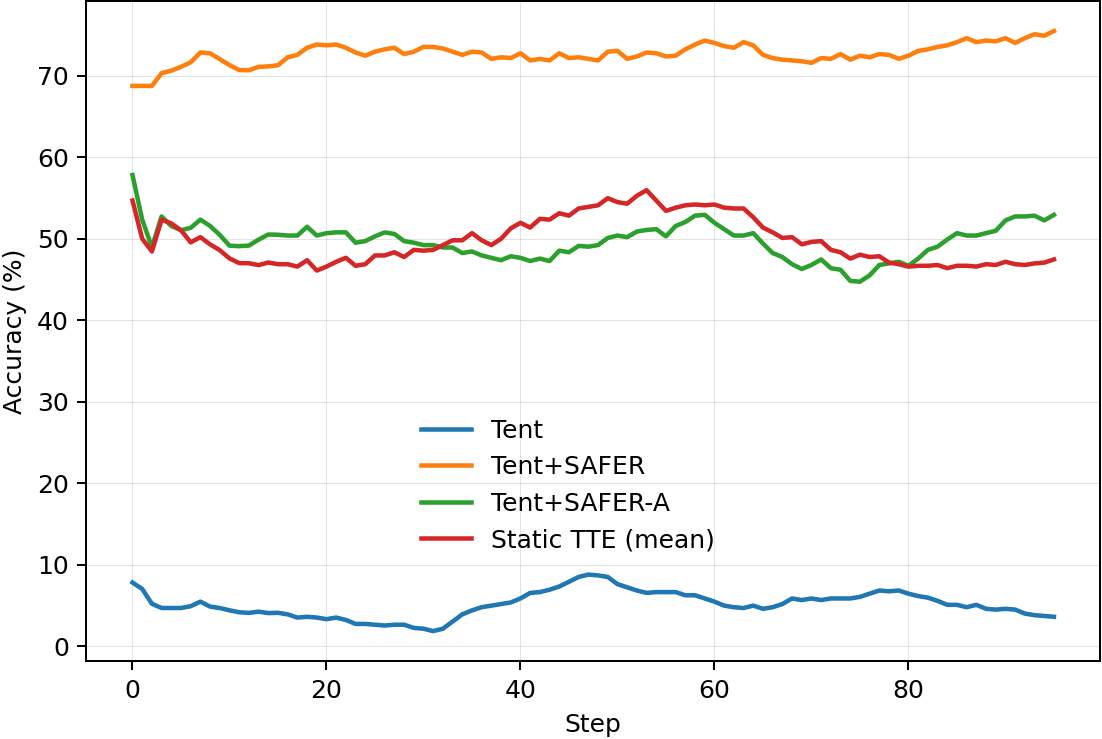}\hfill
\includegraphics[alt={Line plot of accuracy in percent versus attack rate from 0 to 100 percent in 5 percent steps on the PACS Art domain. Curves are shown for Static TTE mean, Tent, Tent+SAFER, and Tent+SAFER-A. Tent falls steeply as the attack rate rises while the SAFER variants degrade much more gradually.},width=0.48\linewidth]{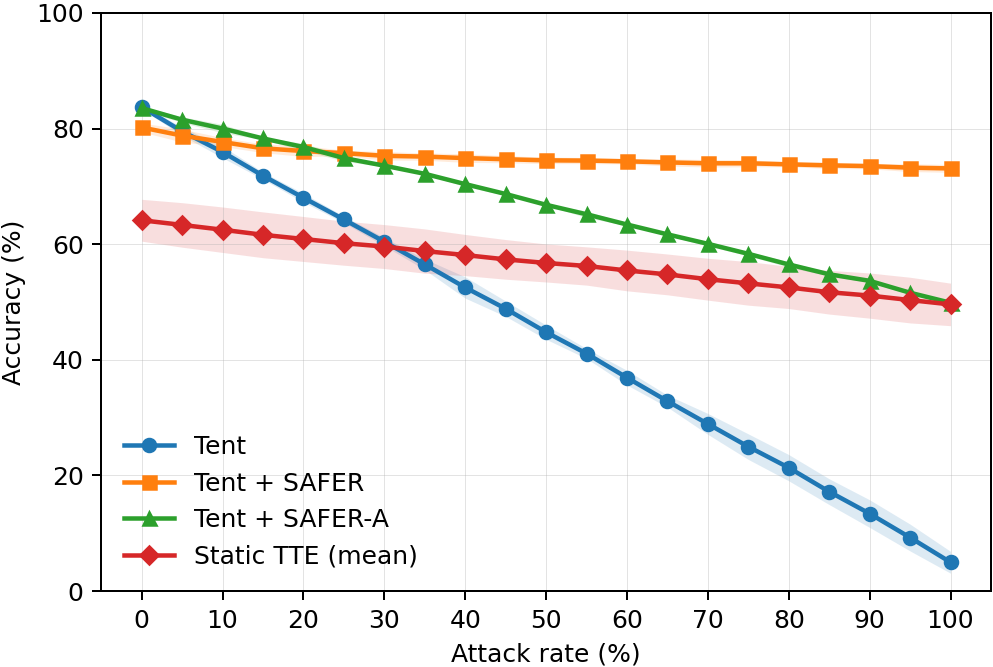}
\caption{Single-domain diagnostics on PACS:Art Painting for Tent-based methods. Left: batch-stability diagnostic using rolling batch accuracy (window 16) for Tent, Tent+SAFER, Tent+SAFER-A, and the no-adaptation Static TTE (mean) reference under a fully attacked ($100\%$) stream. Right: accuracy versus attack rate for the same four methods, over a finer grid (5\% steps).}
\label{fig:abl-stability}
\label{fig:attack-rate-sweep}
\end{figure}

\subsection{Experimental Settings}
\label{sec:experimental-settings}

\paragraph{\bf Datasets.}
Following recent work in TTA \cite{huselective}, we evaluate SAFER on three standard domain generalization datasets: \textbf{PACS} \cite{pacs}, \textbf{VLCS} \cite{vlcs}, and \textbf{OfficeHome} \cite{officehome}. For each dataset, we follow the standard leave-one-domain-out protocol: one domain is held out as the target test domain, and the remaining domains are used to train the source model. At test time, test data arrives as an online stream and each method performs a single sequential pass over the stream without access to source data or target labels. 
Adversarial attacks are added to the test data. 
We report test accuracy results, including both the means and standard deviations over three runs with random seeds.

\paragraph{\bf Attack Setting.}
We evaluate robust test-time adaptation under a black-box adversarial threat model.
We consider an adversary that, for each target domain, injects $\ell_p$-bounded perturbations into a subset of the online test stream at deployment.
We assume the adversary has no access to the deployed model parameters, no gradients through the test-time updates, and no control over the future batch order used by adaptation. The adversary does, however, have access to a labeled dataset drawn from a distribution sufficiently similar to the target domain, on which it trains a surrogate model.
Using this surrogate, adversarially attacked test instances $\tilde{\vx}=\vx+\vepsilon$ are generated offline with iterative projected gradient descent (PGD) \cite{pgd}, where each step updates
\begin{equation}
    \vepsilon \xleftarrow{} \vepsilon + \eta \nabla_{\vepsilon} \gL_{CE}(\vx+\vepsilon, \theta^\star_t)
\end{equation}
followed by projection onto the $\ell_p$ ball. The perturbed instance $\tilde{\vx}$ then replaces $\vx$ in the test stream. We primarily consider an $\ell_\infty$ attack, which enforces $\|\vepsilon\|_\infty \le 8/255$ over 20 PGD steps with step size $\eta=2/255$.
We set these parameters empirically such that the surrogate $\theta^\star_t$ has accuracy near $0\%$ under a $100\%$ attack rate. We pre-generate and store the attacked streams for reproducible comparisons, and all methods are evaluated on identical precomputed streams for a given seed.

\emph{How attacks are injected.} At an attack rate $\gamma\in[0\%,100\%]$, $\gamma$ percent of images per batch are replaced by their adversarial counterpart $\tilde{\vx}$. Intermediate rates therefore yield mini-batches that \emph{mix} clean and adversarial samples. We primarily report $0\%$ (clean) and $100\%$ (fully attacked) in the main tables, with $50\%$ reported in the supplementary material and a finer sweep analyzed in Sec.~\ref{sec:ablations} (Fig.~\ref{fig:attack-rate-sweep}).

\emph{Why black-box transfer.} This is the realistic deployment threat for an online wrapper, and it matches the setting in which TTA-specific adversarial risks were first identified~\cite{wu2023uncovering}. Adaptive white-box attacks that differentiate \emph{through} the evolving adaptation loop are a separate, substantially harder study: such evaluations are prone to gradient-obfuscation pitfalls and must be tailored per defense to be meaningful~\cite{athalye2018obfuscated,tramer2020adaptive,croce2022evaluating}, and stochastic input transforms like ours specifically require Expectation-over-Transformation-style attacks to be assessed soundly~\cite{athalye2018eot}. We therefore scope this study to the transfer setting and treat adaptive attacks as future work. The supplementary material substantiates this scope on PACS:Art: SAFER's protection is essentially unchanged when PGD is replaced by the stronger AutoAttack ensemble at the same budget and degrades gracefully as $\epsilon$ grows to $16/255$, whereas a defense-aware Expectation-over-Transformation attack that differentiates \emph{through} the stochastic augmentations does collapse it, confirming that the gains reported here are specific to the black-box transfer threat model.

\paragraph{\bf Comparison Baselines.}
We compare SAFER against source-only anchors (ERM, Robust ERM), standard TTA baselines (Tent \cite{tent}, PL \cite{pseudolabel}, TSD \cite{tsd}, TeSLA \cite{tesla}), and robustness-oriented baselines (EATA, Tent+MedBN), spanning adaptation strategies based on optimization, pseudo-labels, prototypes/memory, and stability. Robust ERM contextualizes SAFER against an adversarially trained backbone rather than only standard source models, EATA adds explicit sample filtering and update regularization, and Tent+MedBN (Sec.~\ref{sec:related}) is a natural comparator because it likewise defends against malicious test samples in a modular way, by making BN statistic estimation robust during adaptation.

\paragraph{\bf Implementation Details.}
Following prior works~\cite{tsd}, for source domain training, we split the images to $80\%$ for training and $20\%$ for validation. We use the Adam optimizer \cite{adam} with a learning rate of $5e^{-5}$, with weights of source models initialized to a pretrained ImageNet-1K \cite{imagenet1k} baseline. Source model training is done with a batch size of 32, and a maximum of 50 epochs. Dropout probability and weight decay are set to 0. All images are resized to $224\times 224$, and training data is augmented with random cropping, horizontal flipping, color jitter, and intensity changes. For fully online test-time adaptation, we only perform one pass over the test domain, with a batch size of 64, and one adaptation/gradient step per batch. TSD \cite{tsd} and T3A \cite{t3a} both use 100 supports following their literature. The Pseudo-label \cite{pseudolabel} confidence is set to 0.9 following \cite{tsd}. All methods use a ResNet-18 \cite{resnet} backbone with Batch Normalization \cite{batchnorm}. For Tent+MedBN, we use the same Tent adaptation pipeline but replace the standard BN test-batch mean statistic with the MedBN normalization rule during adaptation. For all TTA baselines, as well as the backbone and source domain training code, we use implementations from the released code of the TSD library \cite{tsd}.

\paragraph{\bf SAFER Configuration.}
\label{sec:safer-hyperparams}
Unless otherwise stated, SAFER uses $N=4$ stochastic augmentations in addition to the original view, drawn from the operator family of Sec.~\ref{sec:augmentation} (full definitions and parameter ranges are in the supplementary material). We report two variants (Sec.~\ref{sec:adaptive-alpha}): the default \textbf{SAFER}, which pools over all views directly, and the optional adaptive-mixing \textbf{SAFER-A}. For SAFER-A, we set $\tau=0.2$, $\kappa=3.0$, $\alpha_{\mathrm{atk}}=0.25$, and $\alpha_{\mathrm{clean}}=0.65$. These four SAFER-A values are \emph{fixed across all datasets and wrapped methods}: they were chosen once on a cross-dataset, cross-domain diagnostic, never re-tuned per benchmark. We stress that the default SAFER has essentially \emph{two} knobs (the number of views $N$ and the pooling rule), whereas the additional hyperparameters ($\tau,\kappa,\alpha_{\mathrm{atk}},\alpha_{\mathrm{clean}}$) appear \emph{only} in the optional SAFER-A clean-retention variant. A sensitivity analysis for these parameters is summarized in Sec.~\ref{sec:ablations} and detailed in the supplementary material.

\paragraph{\bf Computational Cost.}
SAFER adds \emph{no new parameters} and \emph{no extra adaptation step}, so the wrapped method still performs a single gradient update per batch, now on the pooled prediction.
The overhead is each batch being expanded into $V = N+1$ stochastic views processed through the \emph{shared} backbone, so both the forward pass and the backpropagation through it grow linearly in $V$; the augmentation operators themselves are inexpensive pixel-space transforms. Fig.~\ref{fig:abl-views-maxops} (left) reports the measured per-step wall-clock time alongside accuracy across $N \in \{1,2,3,4\}$: the measured cost confirms the analytic estimate. Because the largest robustness gains already appear by $N=2$, the approximately $3\times$-compute configuration offers a budget operating point when inference cost is constrained.

\subsection{Experimental Results}

Main black-box transfer results are shown in Tables~\ref{tab:pacs-main}, \ref{tab:vlcs-main}, and \ref{tab:officehome-main}. In the +SAFER rows, SAFER wraps the base TTA method without modifying its native adaptation objective. +SAFER denotes the default reliability-guided pooled predictor, while +SAFER-A denotes the optional adaptive-mixing variant from Sec.~\ref{sec:adaptive-alpha}.

Across datasets, SAFER consistently strengthens robustness under adversarially contaminated streams. While several standard TTA baselines degrade sharply under attack, their SAFER-wrapped versions remain substantially more stable; on PACS and VLCS in particular, Tent+SAFER and TSD+SAFER retain much stronger accuracy under attack than their unwrapped counterparts, indicating that the reliability-guided multi-view predictor materially reduces the vulnerability of online adaptation to adversarial contamination.

SAFER also remains competitive against stronger robustness- or stability-oriented baselines, not only standard TTA methods such as Tent, PL, and TSD. TeSLA is a meaningful comparator because it too exploits augmentation during test-time learning, yet SAFER is often stronger in the adversarial-stream setting, indicating that how views are selected, weighted, and pooled matters, not merely that augmentation is used. Robust ERM provides a non-adaptive robust-source anchor: in some heavily attacked settings it is more stable than fragile online TTA, but SAFER frequently recovers the benefits of adaptation while avoiding much of that collapse.

The relative behavior of SAFER and SAFER-A reveals a consistent tradeoff between robustness and clean accuracy: SAFER usually achieves the highest attacked-stream accuracy, especially at 100\% attack rate, while SAFER-A retains more trust in the original view when disagreement is low, improving clean-stream retention in some settings at the cost of robustness. The clean-stream cost of default SAFER is dataset-dependent: it is typically within one to two points of the unwrapped baseline on PACS and VLCS, but reaches several points on some OfficeHome domains (e.g., Tent on Art and RealWorld), which is precisely the regime SAFER-A is designed to mitigate. That the same wrapper improves multiple underlying TTA methods reinforces SAFER's modularity: it is not tied to one particular adaptation loss but stabilizes several distinct TTA objectives under attacked streams.

\subsection{Ablation Results}
\label{sec:ablations}

All ablations below are conducted on the PACS dataset using the Art Painting domain as the target, unless otherwise stated. We next examine whether the gains of SAFER come from the specific reliability-guided pooling rule and stochastic multi-view design, rather than from merely adding transformed views.

\paragraph{\bf Pooling and Single-Transform Ablation.}
Table~\ref{tab:abl-pooling-tent} compares SAFER against simple single-transform defenses and several view-pooling rules on the same Tent backbone, averaged over PACS domains. JPEG and FFT low-pass filtering, and all pooling variants, substantially improve robustness over the unwrapped baseline, confirming that suppressive transforms and multi-view prediction are each already beneficial in adversarial RTTA. However, full SAFER performs best: the reliability-guided variants are consistently stronger than uniform or entropy-based pooling, especially at high attack rate, and \texttt{cc\_drop} achieves the best 100\% attack-rate performance, exceeding mean pooling, entropy weighting, and every single-transform defense while preserving competitive clean accuracy. The further gain of \texttt{cc\_drop} over \texttt{cc} shows that some views, augmented or original, are actively harmful to include, so lightweight outlier removal is an integral part of the design rather than a cosmetic refinement. SAFER is thus not reducible to a single preprocessing defense or to view multiplicity alone: not all transformed views should contribute equally, and the reliability-guided aggregation itself contributes meaningfully.

\paragraph{\bf Alpha Mode.}
Table~\ref{tab:abl-alpha-mode} isolates the adaptive mixing curve while fixing the rest of the SAFER wrapper. The three adaptive variants differ only in how the disagreement signal $s_b^{\mathrm{feat}}$ is converted into the mixing score $q_b$ used in Eq.~\eqref{eq:alpha-sigmoid}: \emph{Step} applies a hard threshold at $\tau$ (the original view receives $\alpha_{\mathrm{clean}}$ when $s_b^{\mathrm{feat}}<\tau$ and $\alpha_{\mathrm{atk}}$ otherwise), \emph{Linear} replaces the threshold with a clipped linear ramp of fixed width around $\tau$, and \emph{Sigmoid} is the smooth logistic curve of Eq.~\eqref{eq:alpha-sigmoid} itself. The default SAFER setting (no adaptive mixing; the Tent + SAFER row) is strongest at 100\% attack rate (73.10), while all adaptive variants trade robustness for cleaner-stream retention. Among adaptive choices, step and linear are stronger than sigmoid at 100\% attack rate, but still below the no-alpha setting.

\paragraph{\bf Views and Augmentation Budget Sensitivity.}
Figure~\ref{fig:abl-views-maxops} shows two clear trends. First, increasing views from 1 to 4 consistently improves clean and attacked accuracy, with the largest jump from 1 to 2 views. Second, increasing augmentation budget beyond 3 hurts robustness and clean performance, indicating that overly strong augmentation chains can become destructive.

\paragraph{\bf Accuracy versus Attack Rate.}
Fig.~\ref{fig:attack-rate-sweep} (right) traces accuracy across a finer grid of attack rates, adding Static TTE (mean) \cite{perez2021tte} as a no-adaptation reference. The curves span a robustness--plasticity spectrum: unwrapped Tent starts highest but collapses almost linearly to near-chance, dropping below Tent+SAFER almost immediately and below Static TTE by $\gamma\approx25\%$, since adaptation helps most on clean streams but is also what an attacker destabilizes first; Static TTE never updates on the stream, so it starts lowest but decays only mildly. Tent+SAFER dominates nearly the whole sweep, starting close to Tent but degrading so gradually that the gap widens steadily with $\gamma$ rather than only at the $100\%$ extreme. Tent+SAFER-A makes the tradeoff from Sec.~\ref{sec:adaptive-alpha} concrete: it tracks (and briefly exceeds) Tent at low $\gamma$ but degrades faster than Tent+SAFER, converging to the Static TTE/Tent floor by $\gamma=100\%$. SAFER and SAFER-A thus push the same base method to more favorable points on this spectrum, prioritizing robustness and clean-stream retention respectively.

\paragraph{\bf Batch Stability over Time.}
Figure~\ref{fig:abl-stability} (left) reports rolling batch accuracy over the test stream under $100\%$ attack for Tent, Tent+SAFER, Tent+SAFER-A, and Static TTE (mean), the no-adaptation reference. Each adaptation step is one test batch of $64$ samples with a single gradient update (Sec.~\ref{sec:experimental-settings}), and each plotted value is the accuracy averaged over a trailing window of the $16$ most recent batches. Unwrapped Tent remains near collapse throughout (final rolling accuracy $\approx 3.6$), while Tent+SAFER stays high and stable ($\approx 75.5$). Tent+SAFER-A and Static TTE (mean) track each other closely at an intermediate level, finishing around $\approx 53$ and $\approx 47$ respectively, both well above unwrapped Tent but below Tent+SAFER, consistent with the robustness ordering observed across the attack-rate sweep.

\section{Conclusion}

We presented SAFER, a reliability-guided stochastic augmentation ensemble wrapper for robust test-time adaptation on adversarially contaminated streams. SAFER replaces brittle single-view predictions with predictions pooled over the original view and stochastic augmentations, downweights unreliable views through feature agreement, and attaches to existing TTA methods without modifying their native adaptation objectives; the optional SAFER-A extension improves the robustness--clean tradeoff when stronger clean-performance retention is desired. Across PACS, VLCS, and OfficeHome under PGD attacks and varying attack rates, SAFER consistently improves the adversarial robustness of multiple TTA baselines while maintaining competitive clean performance, supporting reliability-guided stochastic ensembling as a practical and broadly compatible design principle for stabilizing online adaptation under attacked target streams.

%%%%%%%%%%%%%%%%%%%%%%%%

\bibliographystyle{splncs04}
\bibliography{refs}

\newpage
\appendix
\section*{Supplementary Materials}

\begin{table*}[t]
\centering
\caption{PACS accuracy (\%) under $\ell_\infty$ attacks with vertically stacked clean/mixed/fully attacked subtables per target domain.}
\label{tab:pacs-main-stacked}
\setlength{\tabcolsep}{3pt}
\textbf{Clean (0\%)}\\
\vspace{0.2em}
\begin{tabular}{l|rrrr}
\hline
Method & Art Painting & Cartoon & Photo & Sketch \\
\hline

ERM & 78.82$_{\pm 2.00}$ & 75.51$_{\pm 2.22}$ & 94.99$_{\pm 0.42}$ & 71.10$_{\pm 1.86}$ \\
Robust ERM & 47.69$_{\pm 0.93}$ & 64.04$_{\pm 1.47}$ & 62.20$_{\pm 0.62}$ & 58.45$_{\pm 1.94}$ \\

\hline
Tent & 83.72$_{\pm 0.35}$ & 83.45$_{\pm 0.79}$ & \underline{95.91}$_{\pm 0.66}$ & 78.31$_{\pm 1.15}$ \\
Tent+MedBN & 82.62$_{\pm 0.93}$ & 81.41$_{\pm 0.47}$ & 95.87$_{\pm 0.59}$ & 75.35$_{\pm 1.05}$ \\
PL & 85.76$_{\pm 2.02}$ & 82.10$_{\pm 3.34}$ & 95.37$_{\pm 0.88}$ & 72.91$_{\pm 5.82}$ \\
EATA & 81.61$_{\pm 0.71}$ & 81.40$_{\pm 0.73}$ & 95.27$_{\pm 0.51}$ & 73.00$_{\pm 0.91}$ \\
TSD & \textbf{86.90}$_{\pm 1.59}$ & \underline{88.10}$_{\pm 0.62}$ & \textbf{96.31}$_{\pm 0.89}$ & 78.90$_{\pm 3.75}$ \\
TeSLA & 82.21$_{\pm 0.91}$ & 80.86$_{\pm 0.42}$ & 95.11$_{\pm 0.09}$ & 74.16$_{\pm 0.60}$ \\

\hline
Tent + SAFER-A & 82.60$_{\pm 1.17}$ & 83.73$_{\pm 0.77}$ & 95.71$_{\pm 0.46}$ & \textbf{81.14}$_{\pm 1.01}$ \\
PL + SAFER-A & 83.63$_{\pm 1.47}$ & 84.33$_{\pm 2.83}$ & 94.43$_{\pm 1.20}$ & 77.76$_{\pm 7.03}$ \\
TSD + SAFER-A & \underline{85.87}$_{\pm 1.89}$ & \textbf{88.41}$_{\pm 0.97}$ & 95.53$_{\pm 1.16}$ & 78.62$_{\pm 1.97}$ \\

\hline
Tent + SAFER & 80.06$_{\pm 1.25}$ & 83.11$_{\pm 0.67}$ & 94.63$_{\pm 0.19}$ & \underline{80.28}$_{\pm 0.92}$ \\
PL + SAFER & 79.13$_{\pm 3.13}$ & 85.00$_{\pm 1.75}$ & 93.93$_{\pm 0.95}$ & 74.73$_{\pm 1.80}$ \\
TSD + SAFER & 83.32$_{\pm 1.20}$ & 87.57$_{\pm 1.22}$ & 94.21$_{\pm 1.05}$ & 76.00$_{\pm 1.04}$ \\

\hline
\end{tabular}

\vspace{0.5em}
\textbf{Mixed (50\%)}\\
\vspace{0.2em}
\begin{tabular}{l|rrrr}
\hline
Method & Art Painting & Cartoon & Photo & Sketch \\
\hline

ERM & 45.26$_{\pm 2.46}$ & 48.66$_{\pm 2.14}$ & 77.43$_{\pm 1.48}$ & 63.16$_{\pm 1.08}$ \\
Robust ERM & 47.54$_{\pm 0.88}$ & 63.84$_{\pm 1.55}$ & 62.18$_{\pm 0.58}$ & 58.39$_{\pm 1.90}$ \\

\hline
Tent & 44.78$_{\pm 1.23}$ & 57.71$_{\pm 1.75}$ & 60.50$_{\pm 2.81}$ & 49.32$_{\pm 6.19}$ \\
Tent+MedBN & 43.73$_{\pm 1.32}$ & 60.00$_{\pm 1.36}$ & 57.21$_{\pm 1.50}$ & 50.75$_{\pm 4.34}$ \\
PL & 45.07$_{\pm 2.08}$ & 58.79$_{\pm 0.66}$ & 64.53$_{\pm 0.86}$ & 44.48$_{\pm 10.24}$ \\
EATA & 43.99$_{\pm 1.31}$ & 57.47$_{\pm 1.70}$ & 60.90$_{\pm 1.94}$ & 47.12$_{\pm 3.68}$ \\
TSD & 44.87$_{\pm 1.10}$ & 65.27$_{\pm 1.13}$ & 59.16$_{\pm 4.45}$ & 47.45$_{\pm 17.77}$ \\
TeSLA & 63.66$_{\pm 1.87}$ & 70.93$_{\pm 0.79}$ & 81.24$_{\pm 2.04}$ & 60.84$_{\pm 2.85}$ \\

\hline
Tent + SAFER-A & 66.50$_{\pm 0.61}$ & 74.15$_{\pm 0.37}$ & 86.41$_{\pm 1.12}$ & 72.78$_{\pm 1.89}$ \\
PL + SAFER-A & 65.77$_{\pm 2.96}$ & 73.18$_{\pm 1.04}$ & 87.15$_{\pm 0.74}$ & 69.52$_{\pm 3.23}$ \\
TSD + SAFER-A & 73.18$_{\pm 1.59}$ & 78.91$_{\pm 1.22}$ & 87.88$_{\pm 1.12}$ & 69.86$_{\pm 3.14}$ \\

\hline
Tent + SAFER & 73.47$_{\pm 0.23}$ & 80.08$_{\pm 0.53}$ & \underline{91.66}$_{\pm 0.35}$ & \textbf{76.68}$_{\pm 0.92}$ \\
PL + SAFER & \underline{74.79}$_{\pm 2.61}$ & \underline{81.74}$_{\pm 0.86}$ & 91.10$_{\pm 0.81}$ & 72.61$_{\pm 2.87}$ \\
TSD + SAFER & \textbf{75.68}$_{\pm 1.39}$ & \textbf{83.86}$_{\pm 1.74}$ & \textbf{92.22}$_{\pm 0.99}$ & \underline{74.30}$_{\pm 1.85}$ \\

\hline
\end{tabular}

\vspace{0.5em}
\textbf{Fully attacked (100\%)}\\
\vspace{0.2em}
\begin{tabular}{l|rrrr}
\hline
Method & Art Painting & Cartoon & Photo & Sketch \\
\hline

ERM & 12.40$_{\pm 3.28}$ & 22.94$_{\pm 3.20}$ & 59.34$_{\pm 3.60}$ & 55.61$_{\pm 1.14}$ \\
Robust ERM & 47.23$_{\pm 0.89}$ & 63.77$_{\pm 1.62}$ & 62.08$_{\pm 0.60}$ & 58.29$_{\pm 1.95}$ \\

\hline
Tent & \phantom{0}4.98$_{\pm 1.89}$ & 35.62$_{\pm 5.95}$ & 40.82$_{\pm 5.20}$ & 29.91$_{\pm 6.33}$ \\
Tent+MedBN & \phantom{0}4.57$_{\pm 1.56}$ & 39.55$_{\pm 2.78}$ & 29.16$_{\pm 5.09}$ & 26.73$_{\pm 7.50}$ \\
PL & \phantom{0}2.78$_{\pm 1.32}$ & 37.54$_{\pm 1.92}$ & 35.19$_{\pm 8.23}$ & 28.34$_{\pm 9.69}$ \\
EATA & \phantom{0}7.39$_{\pm 1.62}$ & 36.08$_{\pm 3.10}$ & 38.00$_{\pm 4.52}$ & 28.54$_{\pm 4.95}$ \\
TSD & \phantom{0}1.61$_{\pm 0.75}$ & 45.05$_{\pm 4.41}$ & 45.63$_{\pm 9.06}$ & 32.72$_{\pm 15.87}$ \\
TeSLA & 45.38$_{\pm 4.63}$ & 61.43$_{\pm 1.09}$ & 71.32$_{\pm 3.28}$ & 52.93$_{\pm 3.90}$ \\

\hline
Tent + SAFER-A & 48.44$_{\pm 1.86}$ & 64.68$_{\pm 0.46}$ & 75.27$_{\pm 1.10}$ & 66.61$_{\pm 2.98}$ \\
PL + SAFER-A & 42.02$_{\pm 2.97}$ & 63.41$_{\pm 3.15}$ & 71.82$_{\pm 12.41}$ & 64.43$_{\pm 6.28}$ \\
TSD + SAFER-A & 56.01$_{\pm 2.93}$ & 64.56$_{\pm 1.44}$ & 79.38$_{\pm 0.33}$ & 64.82$_{\pm 0.90}$ \\

\hline
Tent + SAFER & \underline{71.74}$_{\pm 0.52}$ & 79.24$_{\pm 0.23}$ & \textbf{90.78}$_{\pm 0.87}$ & \textbf{76.33}$_{\pm 0.83}$ \\
PL + SAFER & \textbf{72.97}$_{\pm 1.02}$ & \underline{79.62}$_{\pm 1.99}$ & 89.92$_{\pm 1.29}$ & 70.61$_{\pm 6.43}$ \\
TSD + SAFER & 71.18$_{\pm 3.13}$ & \textbf{82.20}$_{\pm 0.16}$ & \underline{90.12}$_{\pm 0.33}$ & \underline{71.45}$_{\pm 3.86}$ \\

\hline
\end{tabular}
\end{table*}

\begin{table*}[t]
\centering
\caption{VLCS accuracy (\%) under $\ell_\infty$ attacks with vertically stacked clean/mixed/fully attacked subtables per target domain.}
\label{tab:vlcs-main-stacked}
\setlength{\tabcolsep}{3pt}
\textbf{Clean (0\%)}\\
\vspace{0.2em}
\begin{tabular}{l|rrrr}
\hline
Method & Caltech101 & LabelMe & SUN09 & VOC2007 \\
\hline

ERM & \underline{95.27}$_{\pm 1.80}$ & 64.76$_{\pm 0.76}$ & \underline{68.96}$_{\pm 0.82}$ & \underline{72.27}$_{\pm 3.35}$ \\
Robust ERM & 61.51$_{\pm 0.11}$ & 53.61$_{\pm 0.57}$ & 43.70$_{\pm 0.90}$ & 48.68$_{\pm 0.58}$ \\

\hline
Tent & 87.59$_{\pm 3.86}$ & 62.44$_{\pm 0.92}$ & 65.99$_{\pm 1.76}$ & \textbf{72.62}$_{\pm 1.29}$ \\
Tent+MedBN & 85.84$_{\pm 2.43}$ & 62.09$_{\pm 1.00}$ & 64.04$_{\pm 0.81}$ & 71.38$_{\pm 2.54}$ \\
PL & 90.62$_{\pm 6.88}$ & \underline{65.59}$_{\pm 1.66}$ & 65.60$_{\pm 7.99}$ & 68.02$_{\pm 3.16}$ \\
EATA & 82.38$_{\pm 3.02}$ & 59.50$_{\pm 1.32}$ & 61.95$_{\pm 0.98}$ & 71.03$_{\pm 1.57}$ \\
TSD & \textbf{95.64}$_{\pm 2.32}$ & 63.77$_{\pm 0.52}$ & 67.14$_{\pm 1.56}$ & 66.44$_{\pm 6.11}$ \\
TeSLA & 82.54$_{\pm 3.44}$ & 59.60$_{\pm 1.37}$ & 61.84$_{\pm 0.90}$ & 72.02$_{\pm 1.42}$ \\

\hline
Tent + SAFER-A & 87.92$_{\pm 3.59}$ & \textbf{65.89}$_{\pm 0.59}$ & 64.46$_{\pm 0.63}$ & 67.57$_{\pm 0.88}$ \\
PL + SAFER-A & 88.81$_{\pm 6.96}$ & 62.46$_{\pm 2.10}$ & 55.13$_{\pm 6.52}$ & 62.99$_{\pm 0.98}$ \\
TSD + SAFER-A & 94.49$_{\pm 1.66}$ & 63.83$_{\pm 0.71}$ & \textbf{69.37}$_{\pm 1.95}$ & 60.20$_{\pm 2.08}$ \\

\hline
Tent + SAFER & 87.02$_{\pm 4.07}$ & 62.42$_{\pm 0.85}$ & 61.08$_{\pm 1.23}$ & 67.03$_{\pm 0.96}$ \\
PL + SAFER & 87.49$_{\pm 6.25}$ & 61.66$_{\pm 0.64}$ & 50.98$_{\pm 1.08}$ & 64.72$_{\pm 1.98}$ \\
TSD + SAFER & 94.61$_{\pm 1.38}$ & 64.12$_{\pm 0.94}$ & 66.72$_{\pm 2.74}$ & 58.51$_{\pm 3.83}$ \\

\hline
\end{tabular}

\vspace{0.5em}
\textbf{Mixed (50\%)}\\
\vspace{0.2em}
\begin{tabular}{l|rrrr}
\hline
Method & Caltech101 & LabelMe & SUN09 & VOC2007 \\
\hline

ERM & 72.89$_{\pm 1.75}$ & 42.78$_{\pm 2.54}$ & 37.78$_{\pm 1.55}$ & 39.95$_{\pm 1.85}$ \\
Robust ERM & 61.48$_{\pm 0.12}$ & 53.65$_{\pm 0.55}$ & 43.70$_{\pm 0.82}$ & 48.62$_{\pm 0.50}$ \\

\hline
Tent & 60.21$_{\pm 5.42}$ & 42.46$_{\pm 3.94}$ & 38.96$_{\pm 0.69}$ & 46.63$_{\pm 3.71}$ \\
Tent+MedBN & 59.32$_{\pm 2.44}$ & 43.88$_{\pm 3.38}$ & 39.28$_{\pm 0.55}$ & 45.27$_{\pm 2.58}$ \\
PL & 55.88$_{\pm 8.77}$ & 45.38$_{\pm 5.57}$ & 38.22$_{\pm 1.15}$ & 48.96$_{\pm 3.72}$ \\
EATA & 58.35$_{\pm 2.58}$ & 42.14$_{\pm 2.09}$ & 38.38$_{\pm 0.50}$ & 43.97$_{\pm 1.88}$ \\
TSD & 56.44$_{\pm 8.73}$ & 32.91$_{\pm 2.93}$ & 37.00$_{\pm 3.79}$ & 42.83$_{\pm 3.05}$ \\
TeSLA & 69.21$_{\pm 3.50}$ & 50.00$_{\pm 2.58}$ & 48.94$_{\pm 0.63}$ & 58.47$_{\pm 1.94}$ \\

\hline
Tent + SAFER-A & 79.27$_{\pm 2.31}$ & 53.92$_{\pm 3.17}$ & 48.74$_{\pm 2.34}$ & 57.66$_{\pm 1.53}$ \\
PL + SAFER-A & 81.41$_{\pm 4.40}$ & 56.61$_{\pm 2.15}$ & 46.18$_{\pm 2.91}$ & 57.28$_{\pm 1.52}$ \\
TSD + SAFER-A & \underline{87.87}$_{\pm 2.61}$ & \underline{59.15}$_{\pm 1.23}$ & \textbf{54.13}$_{\pm 3.00}$ & 53.54$_{\pm 1.95}$ \\

\hline
Tent + SAFER & 84.41$_{\pm 2.86}$ & 59.06$_{\pm 0.61}$ & \underline{53.43}$_{\pm 0.99}$ & \textbf{62.76}$_{\pm 0.84}$ \\
PL + SAFER & 81.88$_{\pm 4.28}$ & 58.41$_{\pm 1.74}$ & 46.01$_{\pm 1.19}$ & \underline{61.11}$_{\pm 2.10}$ \\
TSD + SAFER & \textbf{93.80}$_{\pm 1.68}$ & \textbf{61.70}$_{\pm 1.16}$ & 50.50$_{\pm 9.77}$ & 51.60$_{\pm 6.48}$ \\

\hline
\end{tabular}

\vspace{0.5em}
\textbf{Fully attacked (100\%)}\\
\vspace{0.2em}
\begin{tabular}{l|rrrr}
\hline
Method & Caltech101 & LabelMe & SUN09 & VOC2007 \\
\hline

ERM & 49.63$_{\pm 3.76}$ & 20.36$_{\pm 4.74}$ & \phantom{0}7.33$_{\pm 3.43}$ & \phantom{0}8.71$_{\pm 4.36}$ \\
Robust ERM & 61.48$_{\pm 0.12}$ & 53.69$_{\pm 0.52}$ & \underline{43.65}$_{\pm 0.78}$ & 48.52$_{\pm 0.54}$ \\

\hline
Tent & 53.80$_{\pm 3.46}$ & 15.95$_{\pm 6.02}$ & \phantom{0}9.18$_{\pm 5.21}$ & 21.45$_{\pm 3.63}$ \\
Tent+MedBN & 50.44$_{\pm 4.33}$ & 19.99$_{\pm 5.75}$ & 11.35$_{\pm 3.42}$ & 20.16$_{\pm 3.08}$ \\
PL & 52.25$_{\pm 5.02}$ & \phantom{0}4.46$_{\pm 1.22}$ & \phantom{0}4.70$_{\pm 2.73}$ & 20.44$_{\pm 5.04}$ \\
EATA & 47.94$_{\pm 1.39}$ & 20.39$_{\pm 3.68}$ & 13.84$_{\pm 3.34}$ & 22.10$_{\pm 2.47}$ \\
TSD & 48.62$_{\pm 7.41}$ & \phantom{0}8.36$_{\pm 5.84}$ & \phantom{0}2.51$_{\pm 1.47}$ & 13.57$_{\pm 3.83}$ \\
TeSLA & 63.86$_{\pm 2.59}$ & 40.51$_{\pm 2.13}$ & 36.25$_{\pm 0.32}$ & 46.26$_{\pm 2.62}$ \\

\hline
Tent + SAFER-A & 73.26$_{\pm 1.83}$ & 42.32$_{\pm 6.23}$ & 33.99$_{\pm 3.97}$ & 48.84$_{\pm 1.46}$ \\
PL + SAFER-A & 72.08$_{\pm 2.29}$ & 45.68$_{\pm 11.73}$ & 36.62$_{\pm 3.83}$ & 51.55$_{\pm 1.04}$ \\
TSD + SAFER-A & 75.81$_{\pm 1.06}$ & 51.51$_{\pm 0.66}$ & 36.23$_{\pm 1.16}$ & 48.75$_{\pm 2.35}$ \\

\hline
Tent + SAFER & 83.67$_{\pm 2.94}$ & \underline{56.76}$_{\pm 0.87}$ & \textbf{48.21}$_{\pm 1.23}$ & \textbf{59.73}$_{\pm 0.18}$ \\
PL + SAFER & \underline{84.31}$_{\pm 5.09}$ & 53.82$_{\pm 1.94}$ & 41.30$_{\pm 2.20}$ & \underline{53.92}$_{\pm 2.45}$ \\
TSD + SAFER & \textbf{84.71}$_{\pm 6.21}$ & \textbf{56.94}$_{\pm 4.91}$ & 42.92$_{\pm 1.47}$ & 48.30$_{\pm 1.14}$ \\

\hline
\end{tabular}
\end{table*}

\begin{table*}[t]
\centering
\caption{OfficeHome accuracy (\%) under $\ell_\infty$ attacks with vertically stacked clean/mixed/fully attacked subtables per target domain.}
\label{tab:officehome-main-stacked}
\setlength{\tabcolsep}{3pt}
\textbf{Clean (0\%)}\\
\vspace{0.2em}
\begin{tabular}{l|rrrr}
\hline
Method & Art & Clipart & Product & RealWorld \\
\hline

ERM & 56.15$_{\pm 0.27}$ & 49.00$_{\pm 0.19}$ & \underline{71.50}$_{\pm 0.62}$ & \textbf{73.77}$_{\pm 0.68}$ \\
Robust ERM & 28.80$_{\pm 1.15}$ & 35.54$_{\pm 0.36}$ & 43.20$_{\pm 0.36}$ & 44.43$_{\pm 0.54}$ \\

\hline
Tent & \textbf{56.43}$_{\pm 1.14}$ & \underline{51.43}$_{\pm 0.34}$ & \textbf{71.70}$_{\pm 0.49}$ & \underline{73.36}$_{\pm 0.52}$ \\
Tent+MedBN & 54.66$_{\pm 1.55}$ & 47.96$_{\pm 0.33}$ & 67.58$_{\pm 0.11}$ & 72.10$_{\pm 0.54}$ \\
PL & 51.22$_{\pm 1.20}$ & 39.12$_{\pm 1.74}$ & 65.80$_{\pm 1.12}$ & 65.70$_{\pm 0.95}$ \\
EATA & 54.95$_{\pm 0.78}$ & 49.60$_{\pm 0.48}$ & 70.50$_{\pm 0.14}$ & 72.91$_{\pm 0.21}$ \\
TSD & \underline{56.32}$_{\pm 0.51}$ & 48.10$_{\pm 1.60}$ & 69.56$_{\pm 1.20}$ & 69.42$_{\pm 0.57}$ \\
TeSLA & 55.49$_{\pm 0.75}$ & 47.28$_{\pm 0.70}$ & 70.26$_{\pm 0.18}$ & 72.61$_{\pm 0.44}$ \\

\hline
Tent + SAFER-A & 54.73$_{\pm 1.11}$ & \textbf{51.83}$_{\pm 0.82}$ & 70.18$_{\pm 0.70}$ & 71.13$_{\pm 0.12}$ \\
PL + SAFER-A & 47.05$_{\pm 0.93}$ & 36.77$_{\pm 5.85}$ & 60.08$_{\pm 0.97}$ & 59.54$_{\pm 2.86}$ \\
TSD + SAFER-A & 54.58$_{\pm 1.09}$ & 46.09$_{\pm 1.40}$ & 66.33$_{\pm 2.06}$ & 65.71$_{\pm 0.86}$ \\

\hline
Tent + SAFER & 51.16$_{\pm 0.87}$ & 51.04$_{\pm 0.71}$ & 67.30$_{\pm 0.72}$ & 68.63$_{\pm 0.31}$ \\
PL + SAFER & 44.14$_{\pm 1.34}$ & 33.47$_{\pm 2.16}$ & 56.60$_{\pm 1.06}$ & 55.95$_{\pm 2.23}$ \\
TSD + SAFER & 51.41$_{\pm 0.99}$ & 44.50$_{\pm 0.93}$ & 63.26$_{\pm 1.89}$ & 63.06$_{\pm 1.14}$ \\

\hline
\end{tabular}

\vspace{0.5em}
\textbf{Mixed (50\%)}\\
\vspace{0.2em}
\begin{tabular}{l|rrrr}
\hline
Method & Art & Clipart & Product & RealWorld \\
\hline

ERM & 28.84$_{\pm 0.07}$ & 28.26$_{\pm 0.78}$ & 42.07$_{\pm 0.37}$ & 38.29$_{\pm 0.44}$ \\
Robust ERM & 28.68$_{\pm 1.10}$ & 35.53$_{\pm 0.39}$ & 43.13$_{\pm 0.35}$ & 44.38$_{\pm 0.49}$ \\

\hline
Tent & 28.38$_{\pm 0.86}$ & 30.64$_{\pm 1.64}$ & 43.34$_{\pm 0.40}$ & 37.53$_{\pm 0.25}$ \\
Tent+MedBN & 27.94$_{\pm 0.62}$ & 30.92$_{\pm 0.81}$ & 44.82$_{\pm 0.80}$ & 37.56$_{\pm 0.21}$ \\
PL & 27.44$_{\pm 1.02}$ & 26.87$_{\pm 1.31}$ & 39.20$_{\pm 0.95}$ & 35.16$_{\pm 0.44}$ \\
EATA & 27.58$_{\pm 0.41}$ & 29.60$_{\pm 1.20}$ & 43.46$_{\pm 0.54}$ & 37.11$_{\pm 0.07}$ \\
TSD & 28.54$_{\pm 0.62}$ & 29.38$_{\pm 1.79}$ & 44.14$_{\pm 2.79}$ & 36.58$_{\pm 0.43}$ \\
TeSLA & 40.30$_{\pm 2.22}$ & 37.48$_{\pm 1.24}$ & \underline{58.21}$_{\pm 1.93}$ & \underline{54.80}$_{\pm 1.20}$ \\

\hline
Tent + SAFER-A & 36.92$_{\pm 1.28}$ & \underline{41.92}$_{\pm 0.76}$ & 57.35$_{\pm 0.31}$ & 54.56$_{\pm 0.25}$ \\
PL + SAFER-A & 32.58$_{\pm 1.49}$ & 28.02$_{\pm 3.56}$ & 47.05$_{\pm 0.06}$ & 42.84$_{\pm 0.96}$ \\
TSD + SAFER-A & 36.79$_{\pm 1.51}$ & 37.72$_{\pm 1.03}$ & 54.70$_{\pm 1.20}$ & 50.03$_{\pm 0.32}$ \\

\hline
Tent + SAFER & \underline{43.50}$_{\pm 0.81}$ & \textbf{47.78}$_{\pm 0.23}$ & \textbf{61.74}$_{\pm 0.61}$ & \textbf{62.21}$_{\pm 0.61}$ \\
PL + SAFER & 34.42$_{\pm 1.59}$ & 31.07$_{\pm 4.25}$ & 47.23$_{\pm 1.61}$ & 49.11$_{\pm 1.06}$ \\
TSD + SAFER & \textbf{43.72}$_{\pm 0.32}$ & 41.28$_{\pm 1.08}$ & 57.01$_{\pm 1.00}$ & 54.72$_{\pm 0.87}$ \\

\hline
\end{tabular}

\vspace{0.5em}
\textbf{Fully attacked (100\%)}\\
\vspace{0.2em}
\begin{tabular}{l|rrrr}
\hline
Method & Art & Clipart & Product & RealWorld \\
\hline

ERM & \phantom{0}2.94$_{\pm 0.37}$ & \phantom{0}8.71$_{\pm 0.94}$ & 13.29$_{\pm 0.47}$ & \phantom{0}4.32$_{\pm 0.26}$ \\
Robust ERM & 28.58$_{\pm 1.18}$ & 35.57$_{\pm 0.37}$ & 43.08$_{\pm 0.25}$ & 44.28$_{\pm 0.53}$ \\

\hline
Tent & \phantom{0}2.62$_{\pm 0.36}$ & 11.66$_{\pm 2.15}$ & 19.61$_{\pm 1.19}$ & \phantom{0}4.04$_{\pm 0.27}$ \\
Tent+MedBN & \phantom{0}2.84$_{\pm 0.39}$ & 14.24$_{\pm 0.69}$ & 21.21$_{\pm 1.56}$ & \phantom{0}5.35$_{\pm 0.48}$ \\
PL & \phantom{0}3.15$_{\pm 0.43}$ & 11.42$_{\pm 0.57}$ & 14.89$_{\pm 4.81}$ & \phantom{0}3.37$_{\pm 0.56}$ \\
EATA & \phantom{0}3.98$_{\pm 0.31}$ & 11.96$_{\pm 1.05}$ & 20.57$_{\pm 1.26}$ & \phantom{0}7.73$_{\pm 0.61}$ \\
TSD & \phantom{0}2.51$_{\pm 0.51}$ & 12.09$_{\pm 2.38}$ & 19.32$_{\pm 1.28}$ & \phantom{0}3.03$_{\pm 0.50}$ \\
TeSLA & 26.15$_{\pm 2.81}$ & 29.37$_{\pm 2.38}$ & 48.36$_{\pm 3.27}$ & 41.06$_{\pm 2.28}$ \\

\hline
Tent + SAFER-A & 19.91$_{\pm 1.23}$ & 32.01$_{\pm 1.49}$ & 44.81$_{\pm 0.77}$ & 38.10$_{\pm 0.76}$ \\
PL + SAFER-A & 15.89$_{\pm 0.65}$ & 20.66$_{\pm 0.94}$ & 29.38$_{\pm 2.27}$ & 20.67$_{\pm 3.59}$ \\
TSD + SAFER-A & 20.31$_{\pm 0.33}$ & 30.44$_{\pm 1.01}$ & 42.73$_{\pm 0.35}$ & 32.64$_{\pm 2.62}$ \\

\hline
Tent + SAFER & \underline{40.26}$_{\pm 0.33}$ & \textbf{46.10}$_{\pm 1.06}$ & \textbf{60.03}$_{\pm 0.45}$ & \textbf{60.55}$_{\pm 0.73}$ \\
PL + SAFER & 30.56$_{\pm 0.31}$ & 30.10$_{\pm 2.91}$ & 45.96$_{\pm 0.32}$ & 44.42$_{\pm 0.69}$ \\
TSD + SAFER & \textbf{40.98}$_{\pm 1.04}$ & \underline{38.06}$_{\pm 1.37}$ & \underline{54.01}$_{\pm 0.68}$ & \underline{48.98}$_{\pm 1.95}$ \\

\hline
\end{tabular}
\end{table*}

\subsection*{Additional SAFER Implementation Details}

\subsubsection*{A. Stochastic augmentation pipeline}
\label{sec:appendix-aug-implementation}

SAFER applies augmentations in pixel space. If an incoming tensor is already normalized, we first invert normalization using the backbone's image statistics, apply augmentations in $[0,1]$, and then re-normalize before forwarding the augmented views through the network. Each stochastic augmentation is sampled from the operator library. The corresponding parameter ranges used in the implementation are:
\begin{center}
\begin{tabular}{ll}
\toprule
\textbf{Operator} & \textbf{Sampled parameters} \\
\midrule
Gaussian blur & kernel size $=9$, $\sigma \sim U(1.0, 3.0)$ \\
Gaussian noise & noise scale $\sigma \sim U(0.01, 0.10)$ \\
FFT low-pass & keep ratio $\sim U(0.2, 0.6)$ \\
Equalize & no continuous parameter \\
Invert & no continuous parameter \\
Solarize & threshold $\sim U(0.1, 0.9)$ \\
Posterize & bits $\in \{2,3,4,5\}$ \\
Contrast & factor $\sim U(0.4, 1.6)$ \\
Brightness & factor $\sim U(0.4, 1.6)$ \\
Saturation & factor $\sim U(0.4, 1.6)$ \\
Sharpness & factor $\sim U(0, 10)$ \\
Shear-X / Shear-Y & angle $\sim U(-20^\circ, 20^\circ)$ \\
Translate-X / Translate-Y & relative shift $\sim U(-0.2, 0.2)$ \\
Rotate & angle $\sim U(-30^\circ, 30^\circ)$ \\
\bottomrule
\end{tabular}
\end{center}

Unless otherwise stated, the reported SAFER runs use the default stochastic sampler with the following constraints:
\begin{itemize}
    \item each sampled pipeline contains at most three operations;
    \item every pipeline is required to include at least one smoothing or frequency-based operator from $\{\texttt{gaussian\_blur},\texttt{fft\_low\_pass}\}$;
    \item operator order is randomized subject to these constraints;
    \item continuous parameters are re-sampled per image, yielding instance-level variation within a batch.
\end{itemize}

In practice, this design encourages each view to preserve coarse semantic content while perturbing appearance in diverse ways. The smoothing/frequency constraint is particularly important for robust test-time adaptation, since it ensures that each sampled view includes at least one transformation intended to suppress high-frequency adversarial artifacts.

\subsubsection*{B. Implemented view-weighting variants}
\label{sec:appendix-view-weighting}

The main paper uses \texttt{cc\_drop} pooling. For completeness, SAFER also implements several alternative view-weighting strategies used in ablations:

\paragraph{Mean Pooling.}
All views receive uniform weight.

\paragraph{Entropy Weighting.}
Each view is weighted by a normalized inverse-entropy score computed from its predictive distribution, so lower-entropy views receive greater mass.

\paragraph{Cross-Correlation Weighting (\texttt{cc}).}
View weights are computed from cross-view feature agreement as defined in Eqs.~(3)--(4) of the main paper, followed by clipping at zero and renormalization, but without removing any view.

\paragraph{Cross-Correlation with Dropping (\texttt{cc\_drop}).}
This is the primary configuration used in the paper. Negative reliabilities are clipped to zero, the least reliable view is removed, and the remaining weights are renormalized as in Eq.~(6) of the main paper.

\paragraph{Top-1 Confidence Weighting.}
A simple confidence-based variant assigns view weights according to normalized top-1 predicted probability.

For the feature-agreement variants (\texttt{cc} and \texttt{cc\_drop}), the weights are shared across the batch because they are estimated from batch-level feature statistics.

\subsubsection*{C. Adaptive mixing implementation}
\label{sec:appendix-alpha-implementation}
\begin{figure}[ht]
\centering
\includegraphics[alt={Line plot of accuracy in percent versus the confidence threshold tau for Tent+SAFER-A with the feature-disagreement signal on the PACS Art domain. Two curves are shown, one for a 0 percent attack rate and one for a 100 percent attack rate.},width=0.48\linewidth]{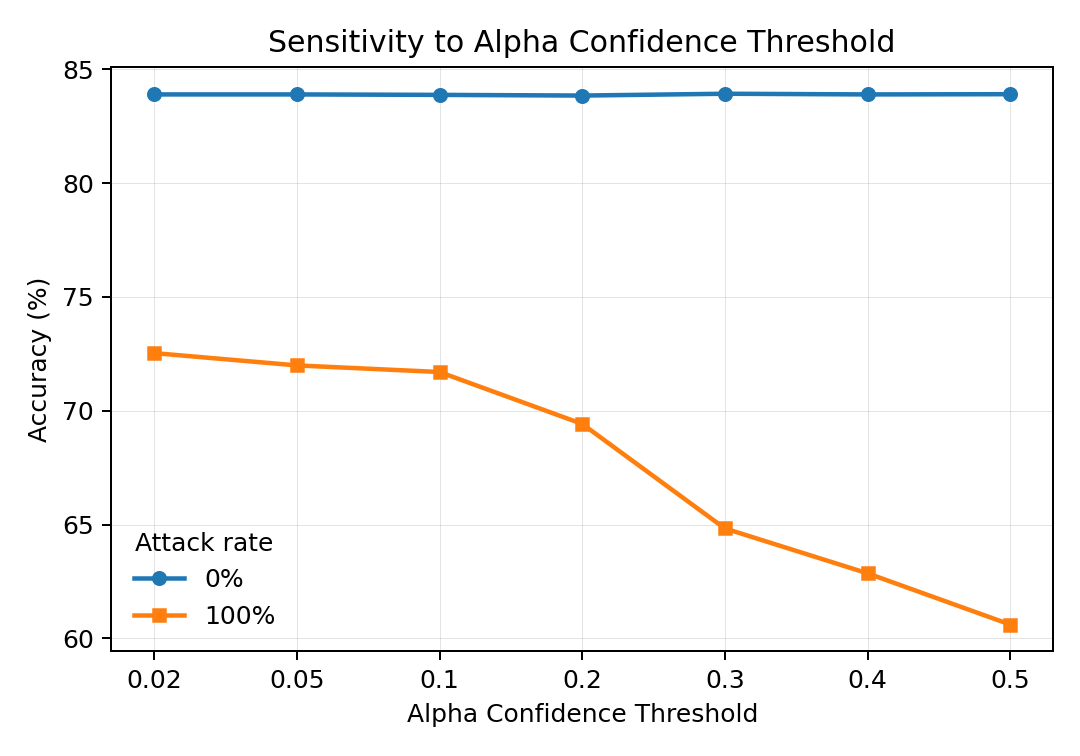}\hfill
\includegraphics[alt={Line plot of accuracy in percent versus the sigmoid slope kappa for Tent+SAFER-A with the feature-disagreement signal on the PACS Art domain. Two curves are shown, one for a 0 percent attack rate and one for a 100 percent attack rate.},width=0.48\linewidth]{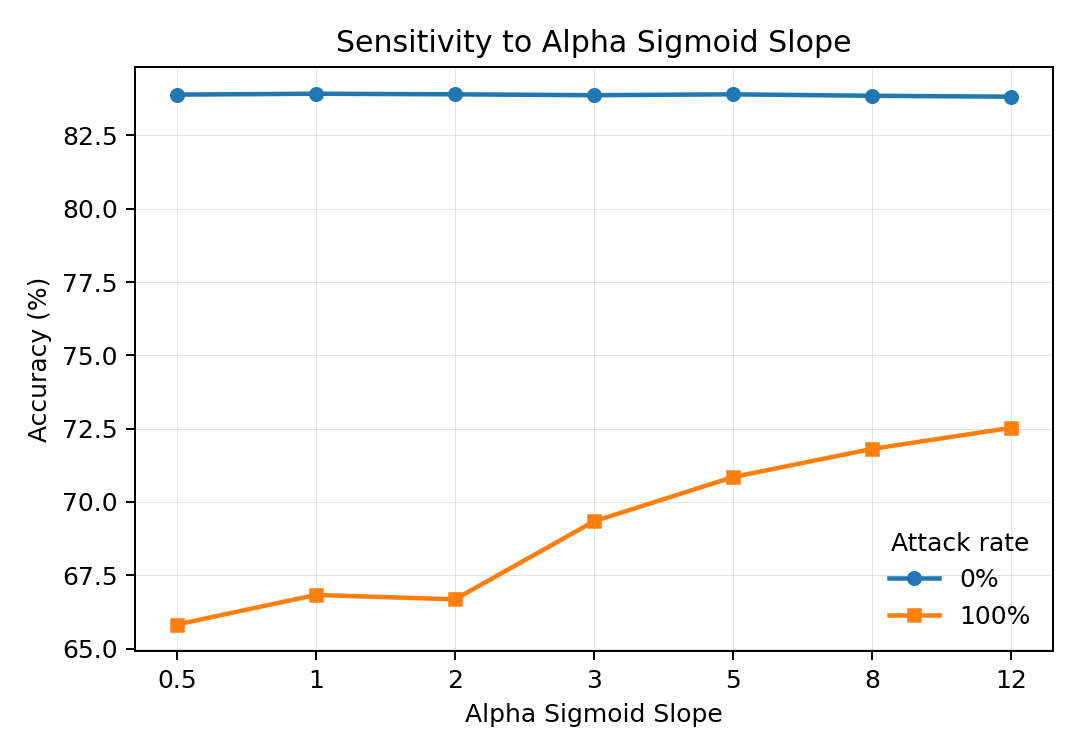}\\[2mm]
\includegraphics[alt={Line plot of accuracy in percent versus the attacked-view mixing value alpha-atk for Tent+SAFER-A with the feature-disagreement signal on the PACS Art domain. Two curves are shown, one for a 0 percent attack rate and one for a 100 percent attack rate.},width=0.48\linewidth]{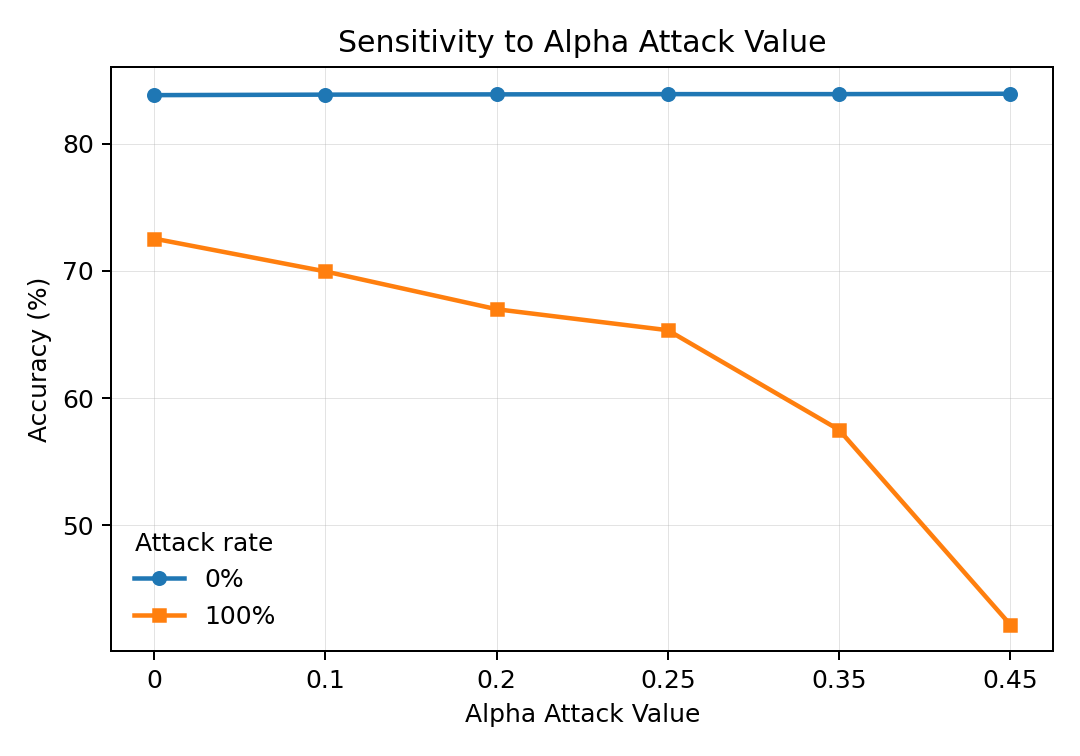}\hfill
\includegraphics[alt={Line plot of accuracy in percent versus the clean-view mixing value alpha-clean for Tent+SAFER-A with the feature-disagreement signal on the PACS Art domain. Two curves are shown, one for a 0 percent attack rate and one for a 100 percent attack rate.},width=0.48\linewidth]{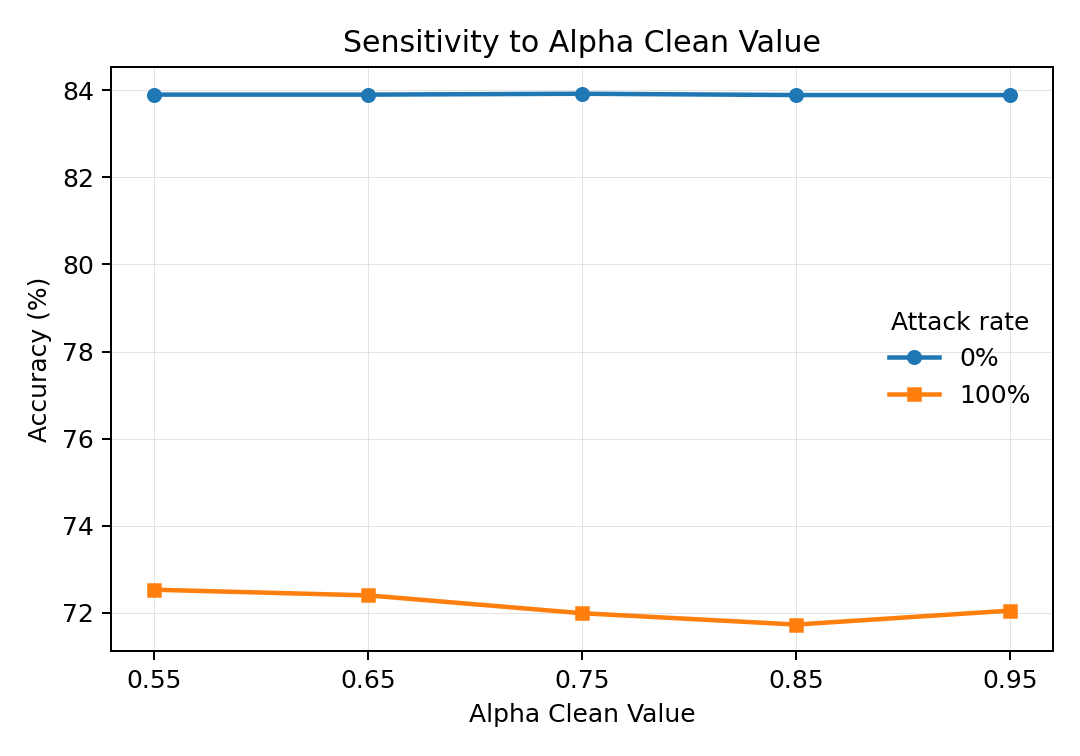}
\caption{Single-domain alpha-parameter sensitivity on PACS:Art Painting (\texttt{Tent+SAFER-A}, \texttt{feat\_disagreement}), with both 0\% and 100\% attack-rate curves shown in each panel. Top-left: threshold $\tau$. Top-right: sigmoid slope $\kappa$. Bottom-left: $\alpha_{\mathrm{atk}}$. Bottom-right: $\alpha_{\mathrm{clean}}$.}
\label{fig:abl-alpha-params}
\end{figure}

\begin{figure}[ht]
\centering
\includegraphics[alt={Two side-by-side heatmaps of Tent+SAFER-A accuracy on the PACS Art domain over a grid of the confidence threshold tau and the sigmoid slope kappa, at a 0 percent attack rate on the left and a 100 percent attack rate on the right. Color encodes accuracy. In the attacked heatmap accuracy is highest when tau is small and kappa is large, and lowest when tau is large with high kappa.},width=0.85\linewidth]{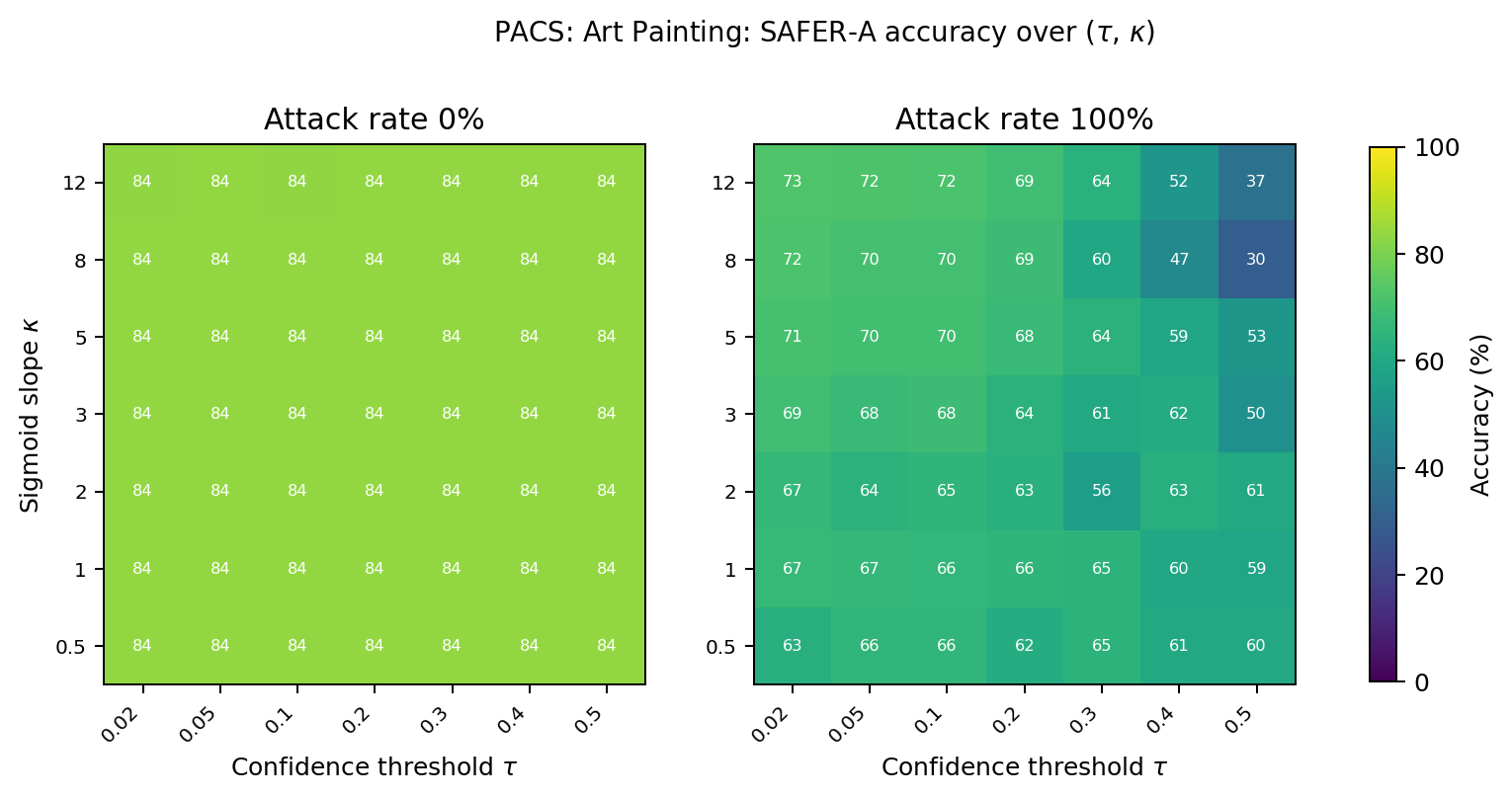}
\caption{Joint $(\tau,\kappa)$ sensitivity of \texttt{Tent+SAFER-A} (\texttt{feat\_disagreement}) on PACS:Art Painting, at $0\%$ (left) and $100\%$ (right) attack rate. Clean accuracy is flat at $84\%$ across the whole grid; the attacked-stream heatmap reveals an interaction the one-dimensional sweeps in Fig.~\ref{fig:abl-alpha-params} cannot show. $\kappa$ acts as a multiplier on whatever $\tau$ already does: a high $\kappa$ is an asset when $\tau$ is small (best cell in the grid: $73\%$ at $\tau=0.02,\kappa=12$) but the single most damaging combination when $\tau$ is large ($30$--$37\%$ at $\tau\in\{0.4,0.5\}$ with $\kappa\in\{8,12\}$, the worst region of the entire grid). At low $\kappa$ ($\le 2$), accuracy is comparatively insensitive to $\tau$, staying in a narrower $56$--$67\%$ band.}
\label{fig:supp-alpha-heatmap}
\end{figure}

The main paper reports two prediction modes: \textbf{SAFER}, which uses the pooled predictor directly, and \textbf{SAFER-A}, which additionally applies adaptive mixing with the original view. Adaptive mixing affects only the final prediction used at inference; it does not alter the wrapped TTA objective itself.

The main SAFER-A setting uses the feature-disagreement signal from Eq.~(7) of the main paper. In implementation, cosine similarity is computed on flattened features and numerically clamped to a valid range before forming the disagreement score. The resulting signal is then mapped to an original-view weight using Eq.~(8) of the main paper, with
\begin{equation}
\label{eq:appendix-alpha-config}
\tau=0.2,\qquad
\kappa=3.0,\qquad
\alpha_{\mathrm{atk}}=0.25,\qquad
\alpha_{\mathrm{clean}}=0.65.
\end{equation}
Thus, larger disagreement pushes the original-view weight toward $\alpha_{\mathrm{atk}}$, while smaller disagreement keeps it closer to $\alpha_{\mathrm{clean}}$.

Beyond the reported sigmoid setting, the implementation also supports two simpler families for ablation:
\begin{equation}
\label{eq:appendix-alpha-oriented}
o_b=
\begin{cases}
s_b-\tau, & \text{if larger values of } s_b \text{ indicate greater suspicion},\\
\tau-s_b, & \text{otherwise},
\end{cases}
\end{equation}
and then
\begin{equation}
\label{eq:appendix-alpha-families}
q_b=
\begin{cases}
1, & \text{step and } o_b \ge 0,\\
0, & \text{step and } o_b < 0,\\
\mathrm{clip}\!\left(0.5 + \frac{o_b}{\delta},\, 0,\, 1\right), & \text{linear},\\
\sigma(\kappa o_b), & \text{sigmoid}.
\end{cases}
\end{equation}
The experiments in the main paper use only the sigmoid family.

The implementation also exposes several candidate mixing signals for
ablation:
\begin{itemize}
\item confidence- and margin-based: \texttt{orig\_conf},
  \texttt{orig\_margin}, \texttt{aug\_margin};
\item entropy-based: \texttt{orig\_entropy}, \texttt{aug\_entropy};
\item gap-based: \texttt{margin\_gap}, \texttt{entropy\_gap};
\item disagreement-based: \texttt{prob\_disagreement},
  \texttt{feat\_disagreement}.
\end{itemize}
Among these, \texttt{feat\_disagreement} is used in all reported
SAFER-A results.

\subsubsection*{D. Wrapper runtime configuration}
\label{sec:appendix-wrapper-config}

For the wrapper experiments reported in the main paper, SAFER uses the following runtime setting unless otherwise stated:
\begin{equation}
\label{eq:appendix-wrapper-defaults}
\begin{aligned}
&N=4,\qquad
\texttt{s\_primary\_view\_pool}=\texttt{cc\_drop},\\
&\texttt{s\_alpha\_mode}\in\{\texttt{none},\texttt{sigmoid}\}.
\end{aligned}
\end{equation}
Here, setting \texttt{s\_alpha\_mode} to \texttt{none} corresponds to
\textbf{SAFER}, while setting it to \texttt{sigmoid} corresponds to
\textbf{SAFER-A}. In both cases, the wrapped TTA method retains its
original optimization rule and update parameters; SAFER only changes
the predictor supplied to that method.

\paragraph{Alpha-Parameter Sensitivity.}
Figure~\ref{fig:abl-alpha-params} shows one-dimensional sweeps with both 0\% (clean) and 100\% attack-rate curves. Clean accuracy is almost flat across all tested alpha hyperparameters (about 83.8--83.9). In contrast, the 100\% curves are sensitive: in isolation, lower $\tau$ and higher $\kappa$ look better, increasing $\alpha_{\mathrm{atk}}$ sharply hurts robustness (more trust in the original view under attack), and sensitivity to $\alpha_{\mathrm{clean}}$ is comparatively mild. However, $\tau$ and $\kappa$ do not act independently: Figure~\ref{fig:supp-alpha-heatmap} sweeps them jointly and shows that $\kappa$ is better read as a multiplier on $\tau$'s effect rather than as an independently "good" knob: a high $\kappa$ is the best choice when $\tau$ is small, but the worst choice when $\tau$ is large, while low $\kappa$ leaves accuracy comparatively insensitive to $\tau$.

\subsubsection*{E. What is fixed across reported SAFER runs}
\label{sec:appendix-fixed-settings}

To make the reported comparisons interpretable, the following SAFER design choices are fixed across the main wrapper experiments:
\begin{itemize}
    \item four stochastic augmentations in addition to the original view;
    \item the same augmentation library and parameter ranges across datasets and wrapped methods;
    \item \texttt{cc\_drop} as the primary reliability-guided pooling rule;
    \item one forward pass per sampled view, with shared backbone parameters;
    \item no source data, no source labels, and no additional training beyond the wrapped TTA procedure.
\end{itemize}

This separation is deliberate: the main claim of the paper is that the same SAFER wrapper can be attached to different TTA objectives with minimal modification, rather than that each wrapped method requires a heavily customized defense.

\subsection*{Additional Threat Models and Robustness Checks}

\paragraph{AutoAttack Transfer Set.}
To verify that SAFER's gains are not specific to PGD or to surrogate-side gradient masking, we regenerate the surrogate's attacked stream with AutoAttack (Croce \& Hein, ICML 2020), the standard parameter-free ensemble of APGD-CE, APGD-T, FAB-T, and the black-box Square attack, at the same $\ell_\infty$ budget $\epsilon=8/255$ on a single domain (PACS:Art), and compare \texttt{Tent} against \texttt{Tent+SAFER} at a $100\%$ attack rate. This experiment deliberately keeps the same black-box \emph{transfer} threat model as the main paper: the perturbations are crafted offline on the independent surrogate and replayed into the test stream, and the deployed adaptation model is never differentiated through. It therefore isolates a single question, namely whether the main-paper result is an artifact of PGD as an attack family, and is complementary to the adaptive, defense-aware setting studied separately below (where the attacker instead targets the deployed SAFER predictor directly). At this budget AutoAttack drives the surrogate to $0\%$ accuracy (APGD-CE alone flips every surrogate prediction), so the resulting stream is at least as strong as the $20$-step PGD stream used elsewhere.

\begin{table}[ht]
\centering
\caption{AutoAttack transfer set on PACS:Art at $\epsilon=8/255$ and a $100\%$ attack rate. Accuracy ($\%$, mean $\pm$ std over seeds $\{0,1,2\}$). Both columns use the identical black-box transfer protocol; only the surrogate-side attack family differs ($20$-step PGD vs.\ the AutoAttack standard ensemble). \texttt{Tent+SAFER} is essentially unchanged between the two attacks, whereas unwrapped \texttt{Tent} collapses further under the stronger AutoAttack.}
\label{tab:supp-autoattack}
\begin{tabular}{l|rr}
\hline
Method & PGD ($8/255$) & AutoAttack ($8/255$) \\
\hline
\texttt{Tent} & 4.98 $\pm$ 1.89 & 2.69 $\pm$ 0.77 \\
\texttt{Tent+SAFER} & 73.10 $\pm$ 0.68 & 73.31 $\pm$ 0.92 \\
\hline
\end{tabular}
\end{table}

The result confirms that SAFER's robustness is not specific to PGD.
Unwrapped \texttt{Tent} is already near collapse under transfer PGD
(at $4.98\%$) and falls even lower under the stronger AutoAttack
stream (to $2.69\%$), consistent with AutoAttack being the harder
attack. By contrast, \texttt{Tent+SAFER} is statistically
indistinguishable across the two attacks: $73.10\%$ versus $73.31\%$,
a gap well within the seed-to-seed deviation. Swapping the attack
family for a stronger, qualitatively different one thus leaves the
wrapper's protection intact. This is expected under the transfer
threat model: SAFER's reliability-guided cross-view pooling suppresses
whichever views a transferred perturbation has corrupted, and that
mechanism is largely agnostic to how the perturbation was optimized on
the surrogate.

\paragraph{Perturbation-Budget Sweep.}
We further sweep the attack budget $\epsilon$ over the values
$2/255$, $4/255$, $8/255$, and $16/255$ on a single domain, comparing
\texttt{Tent} and \texttt{Tent+SAFER}, to characterize graceful
degradation as the perturbation strength grows.
\begin{figure}[ht]
\centering
\includegraphics[alt={Line plot of accuracy in percent versus the L-infinity perturbation budget epsilon, at 2, 4, 8, and 16 over 255, on the PACS Art domain at a 100 percent attack rate, comparing Tent and Tent+SAFER. Tent collapses to near zero even at the smallest budget while Tent+SAFER stays above 65 percent across the whole range.},width=0.6\linewidth]{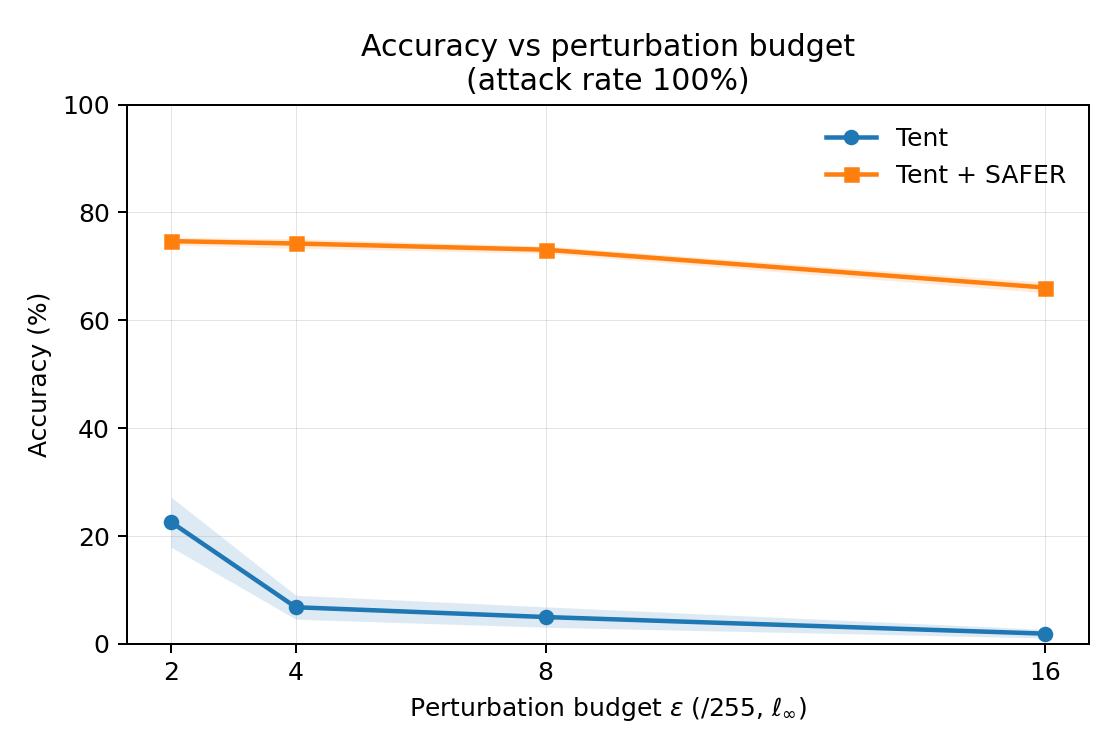}
\caption{Accuracy versus $\ell_\infty$ budget $\epsilon\in\{2,4,8,16\}/255$ on PACS:Art at a $100\%$ attack rate, for \texttt{Tent} vs \texttt{Tent+SAFER}. \texttt{Tent} collapses already at $\epsilon=2/255$ (from $\sim$23\% down to $\sim$2\% at $\epsilon=16/255$), whereas \texttt{Tent+SAFER} degrades gracefully, staying above $65\%$ across the full sweep, confirming that SAFER's gains hold across perturbation strengths rather than at a single tuned $\epsilon$.}
\label{fig:supp-eps-sweep}
\end{figure}

\paragraph{Defense-Aware (Adaptive) Attack.}
As a stress test, we consider an attacker aware of SAFER's stochastic augmentations and explicitly designed to defeat them: an Expectation-over-Transformation (EOT) attack (Athalye et al., ICML 2018), which estimates the PGD gradient by averaging over $K$ freshly sampled SAFER views per step rather than attacking a single (surrogate) view. Even a partial or negative result here is informative, since it bounds SAFER's robustness under a defense-aware adversary and directly addresses the ``what if the attacker knows about SAFER?'' question.

\begin{figure}[ht]
\centering
\includegraphics[alt={Line plot of accuracy in percent versus the number of EOT samples per PGD step K, at 0, 1, 4, and 8, on the PACS Art domain at L-infinity budget 8/255 and a 100 percent attack rate. Curves are shown for Tent+SAFER, Tent+SAFER-A, and a mean-pooling Tent+SAFER variant, with the unwrapped Tent accuracy drawn as a dashed floor. The SAFER variants drop sharply once the attacker uses one or more EOT samples.},width=0.6\linewidth]{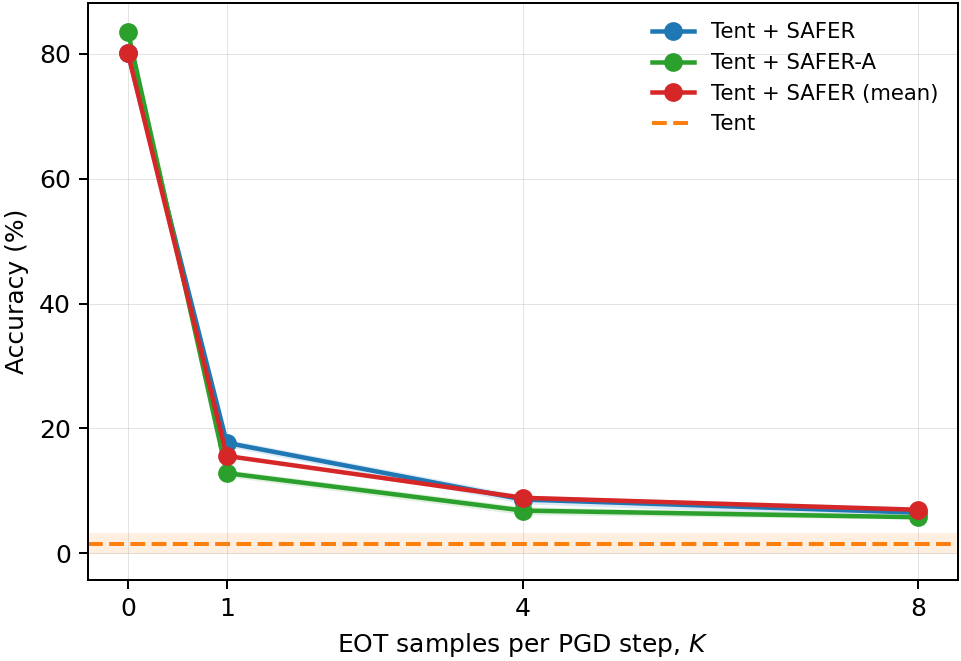}
\caption{Accuracy on PACS:Art at $\epsilon=8/255$ and a $100\%$ attack rate as a function of the number of EOT samples per PGD step, $K\in\{0,1,4,8\}$ ($K=0$ is the standard transfer-PGD attack used elsewhere in the paper), for \texttt{Tent+SAFER}, \texttt{Tent+SAFER-A}, and a mean-pooling \texttt{Tent+SAFER} variant, against the unwrapped \texttt{Tent} floor (dashed).}
\label{fig:supp-eot-attack}
\end{figure}

This is a genuine negative result for the defense-aware setting: a single EOT sample already collapses all three SAFER variants from $\sim$80--84\% at $K=0$ to $\sim$13--18\%, and the accuracy continues to erode toward the unwrapped \texttt{Tent} floor ($\sim$1--2\%) as $K$ grows, reaching $\sim$6--9\% at $K=4$ and $\sim$6--7\% at $K=8$, a roughly tenfold drop from the black-box transfer numbers reported in the main paper. Pooling strategy makes little difference under this attack: \texttt{SAFER}, \texttt{SAFER-A}, and mean pooling all converge to essentially the same curve, indicating that the vulnerability lies in the stochastic-augmentation mechanism itself (which EOT directly targets by averaging it out of the gradient) rather than in any particular reliability-weighting choice. A small residual margin over unwrapped \texttt{Tent} persists even at $K=8$, but it is far smaller than the gains SAFER provides against transfer-based attacks. These results confirm that SAFER's robustness guarantees are scoped to the black-box transfer threat model studied in the main paper, and that a fully adaptive, defense-aware adversary remains an open and important challenge for stochastic-augmentation defenses in RTTA.

\end{document}